\documentclass{article}

\PassOptionsToPackage{numbers}{natbib}
\usepackage[final]{neurips_2021}

\usepackage[utf8]{inputenc} % allow utf-8 input
\usepackage[T1]{fontenc}    % use 8-bit T1 fonts
\usepackage[colorlinks]{hyperref}       % hyperlinks
\usepackage{url}            % simple URL typesetting
\usepackage{booktabs}       % professional-quality tables
\usepackage{amsfonts}       % blackboard math symbols
\usepackage{nicefrac}       % compact symbols for 1/2, etc.
\usepackage{microtype}      % microtypography
\usepackage{xcolor}         % colors

\usepackage{mathtools}
\usepackage{amssymb}
\usepackage{amsthm}
\usepackage{amsfonts}
\usepackage{bm}
\usepackage{bbm}

\usepackage{adjustbox}
\usepackage{multirow}
\usepackage{subfigure}
\usepackage{wrapfig}
\usepackage{placeins}

\usepackage{algorithm}
\usepackage{algorithmic}

\newtheorem{theorem}{Theorem}

\newtheorem{lemma}[theorem]{Lemma}

\definecolor{BrickRed}{rgb}{0.6,0,0}
\definecolor{RoyalBlue}{rgb}{0,0,0.8}
\definecolor{Tdgreen}{rgb}{0,0.4,0.7}

\hypersetup{citecolor=RoyalBlue}

\newcommand{\jh}[1]{{{#1}}}

\newcommand{\ie}{\textit{i}.\textit{e}., }
\newcommand{\eg}{\textit{e}.\textit{g}., }

\newcommand{\pz}{\phantom{0}}
\newcommand{\pms}[1]{\tiny{$\pm$#1}}

\usepackage{pifont}% http://ctan.org/pkg/pifont

\DeclareMathOperator*{\argmax}{arg\,max}

\renewcommand{\algorithmicrequire}{\textbf{Input:}}
\renewcommand{\algorithmicensure}{\textbf{Output:}}

\title{SmoothMix: Training Confidence-calibrated Smoothed Classifiers for Certified Robustness}

\author{
  Jongheon Jeong\textsuperscript{\normalfont 1} \qquad Sejun Park\textsuperscript{\normalfont 2}\thanks{Work done at KAIST.} \qquad Minkyu Kim\textsuperscript{\normalfont 3}
  \\ \\
  \textbf{Heung-Chang Lee}\textsuperscript{\normalfont 4} \qquad \textbf{Doguk Kim}\textsuperscript{\normalfont 5}\thanks{Work done at Kakao Enterprise.} \qquad \textbf{Jinwoo Shin}\textsuperscript{\normalfont 3,1}
  \\ \\ 
  \textsuperscript{1}School of Electrical Engineering, KAIST \quad \textsuperscript{2}Vector Institute for Artificial Intelligence\\
  \textsuperscript{3}Kim Jaechul Graduate School of AI, KAIST \quad \textsuperscript{4}Kakao Enterprise\\
  \quad \textsuperscript{5}Department~of Artificial Intelligence, Inha University\\
  \texttt{\{jongheonj,\,minkyu.kim,\,jinwoos\}@kaist.ac.kr}\\
  \texttt{sejun.park@vectorinstitute.ai}\\
  \texttt{andrew.com@kakaoenterprise.com} \quad \texttt{dgkim@inha.ac.kr}\\
}

\begin{document}

\maketitle

\begin{abstract}
\emph{Randomized smoothing} is currently a state-of-the-art method to construct a \emph{certifiably robust} classifier from neural networks against $\ell_2$-adversarial perturbations. Under the paradigm, the robustness of a classifier is aligned with the \emph{prediction confidence}, \ie the higher confidence from a smoothed classifier implies the better robustness. This motivates us to rethink the fundamental trade-off between accuracy and robustness in terms of \emph{calibrating} confidences of a smoothed classifier. In this paper, we propose a simple training scheme, coined \emph{SmoothMix}, to control the robustness of smoothed classifiers via \emph{self-mixup}: it trains on convex combinations of samples along the direction of adversarial perturbation for each input. The proposed procedure effectively identifies over-confident, near off-class samples as a cause of limited robustness in case of smoothed classifiers, and offers an intuitive way to adaptively set a new decision boundary between these samples for better robustness. Our experimental results demonstrate that the proposed method can significantly improve the certified $\ell_2$-robustness of smoothed classifiers compared to existing state-of-the-art robust training methods.\footnote{Code is available at \url{https://github.com/jh-jeong/smoothmix}.}
\end{abstract}

\section{Introduction}
\label{s:intro}
\emph{Adversarial examples} \cite{szegedy2013intriguing, goodfellow2014explaining} in deep neural networks clearly highlight that neural networks often generalize differently from humans, at least without an additional prior of \emph{local smoothness} of predictions with respect to the input space: an adversarially-crafted, yet imperceptible input perturbation can drastically change the prediction of a neural network based classifier.
Due to the intrinsic complexity of neural networks, however, the community has noticed that it is extremely hard to directly encode this smoothness prior into neural networks \cite{carlini2017adversarial, pmlr-v80-athalye18a, tramer2020adaptive}, especially without relying on \emph{adversarial training} \cite{madry2018towards, pmlr-v97-zhang19p}, \ie augmenting training data with its adversarial examples. 
Even with adversarial training, (a) the non-convex, minimax nature of the training introduces many optimization difficulties, often resulting in a harsh over-fitting \cite{schmidt2018adversarially, pmlr-v119-rice20a}, and (b) it is generally hard to provably guarantee that the learned classifier is indeed smooth. 

\emph{Randomized smoothing} \cite{lecuyer2019certified, pmlr-v97-cohen19c} is relatively a recent idea that aims to \emph{indirectly} encode the smoothness prior: \citet{pmlr-v97-cohen19c} have shown that any classifier, regardless of whether it is smooth or not, can be transformed into a \emph{certifiably robust} classifier via averaging its predictions over Gaussian noise, where the (certified) robustness of the classifier depends on how well the base classifier performs with the noise. Compared to adversarial training, this notion of ``indirect'' smoothness can be favorable in a sense that (a) it is easier to optimize, and (b) offers a provable guarantee on the robustness. Currently, randomized smoothing is considered as the state-of-the-art approach in the context of building a neural network based classifier that is certifiably robust on $\ell_2$-perturbations \cite{li2020sokcertified}.

In this respect, a growing body of research has focused on improving the robustness guarantee that randomized smoothing can give, \eg via different smoothing measures \cite{lee2019tight, pmlr-v119-yang20c} or improved certification procedures \cite{dvijotham2020a, mohapatra2020higher}. 
One of important directions on this line of research is to investigate \emph{which training} of the base classifier could maximize the certified robustness of smoothed classifiers \cite{nips_salman19, Zhai2020MACER, jeong2020consistency}. In particular, \citet{nips_salman19} have proposed \emph{SmoothAdv}, showing that employing adversarial training \cite{madry2018towards} for smoothed classifiers could further improve the robustness, akin to the standard neural networks.
This motivates us to develop a new form of adversarial training, more specialized for smoothed classifiers.

\begin{figure*}[t]
	\centering
	\hfill
	\subfigure[Adversarial training \cite{madry2018towards}]
	{
	    \includegraphics[height=0.9in]{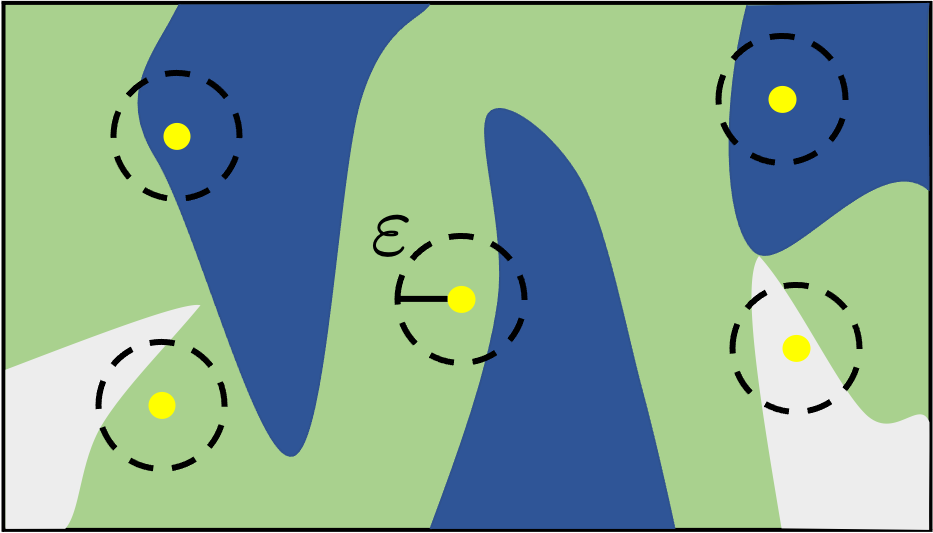}
		\label{fig:at}
	}
	\hfill
	\subfigure[SmoothAdv \cite{nips_salman19}]
	{
	    \includegraphics[height=0.9in]{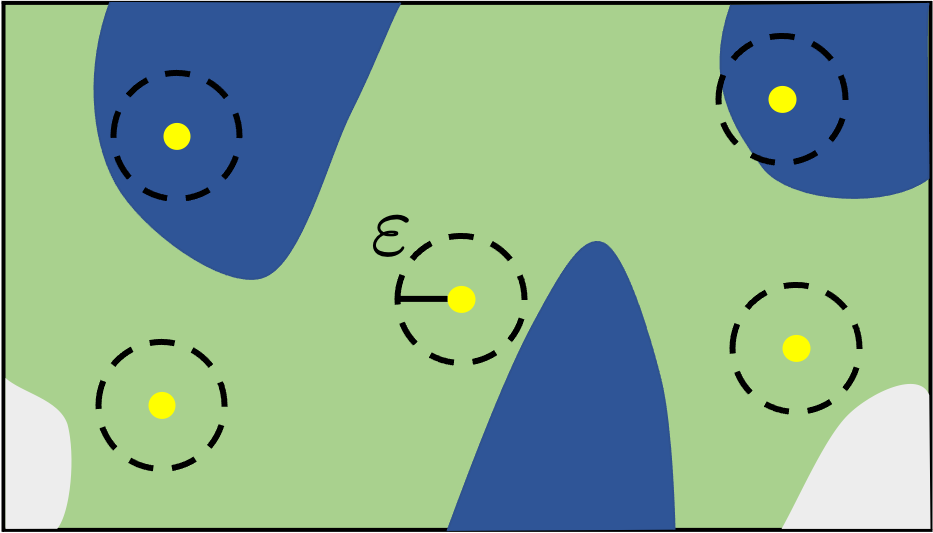}
		\label{fig:smoothadv}
	}
	\hfill
	\subfigure[SmoothMix (Ours)]
	{
	    \includegraphics[height=0.9in]{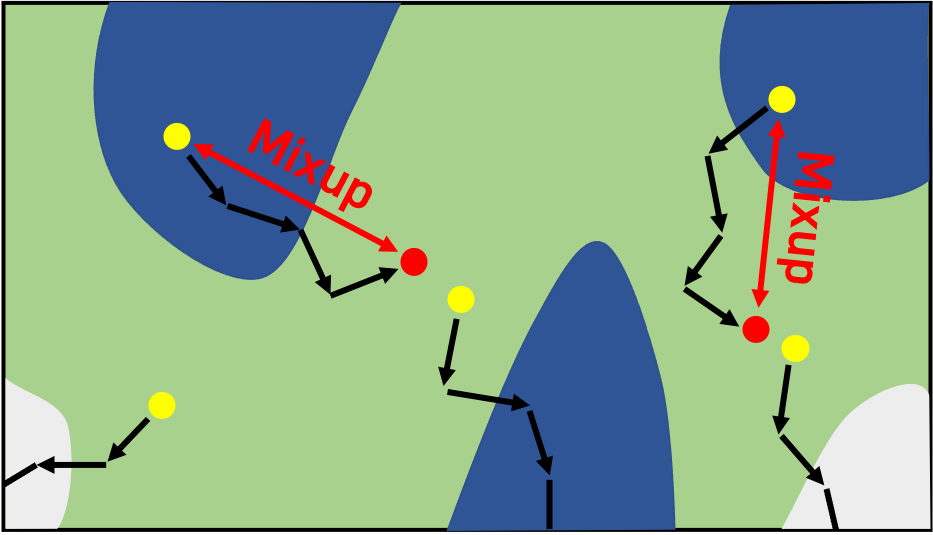}
		\label{fig:smoothmix}
	}
	\hfill
	\caption{Illustrations of how each training method obtains adversarial robustness: adversarial training \cite{madry2018towards} considers an $\varepsilon$-ball around each sample and corrects adversarial examples found in these balls; SmoothAdv \cite{nips_salman19} directly employs adversarial training on smoothed classifiers; SmoothMix (ours) can be differentiated from SmoothAdv as it (\textit{i}) does not assume an explicit norm restriction on adversarial examples, and (\textit{ii}) applies \textit{mixup} \cite{zhang2018mixup} instead of correcting the adversarial examples.}
	\label{fig:concept}
    \vspace{-0.1in}
\end{figure*}

\textbf{Contribution. }
In this paper, we propose \emph{SmoothMix}, a novel adversarial training method designed for improving the certified robustness of smoothed classifiers.
One of the key features that smoothed classifiers offer is a direct correspondence from \emph{prediction confidence} to adversarial robustness: achieving a higher confidence in a smoothed classifier implies that the classifier can give a better certified robustness.
Inspired by this, we found that the certified robustness of a given data sample can be significantly decreased by nearby \emph{off-class} but \emph{over-confident} \cite{pereyra2017regularizing} inputs: such ``harmful'' inputs would occupy an unnecessarily large robust radius near the sample of our interest, especially when they do not contain much information to be discriminated by the classifier.

Under the finding, we aim to {calibrate} the confidence of these off-class inputs to improve the certified robustness at the original input. More specifically, we first observe that such over-confident examples can be efficiently found along the direction of \emph{adversarial} perturbations for a given input. 
Then, we suggest to regularize the over-confident predictions along the adversarial direction toward the \emph{uniform} prediction through a \emph{mixup} loss \cite{zhang2018mixup} (see Figure~\ref{fig:concept} for an overview). 
This new approach of incorporating adversarial examples effectively permits more distant examples in training, even when they go off-class, based on the local-smoothness of smoothed classifiers. It also suggests an intuitive way of defining confidence beyond the given data samples to smoothed classifiers. 

We evaluate our proposed SmoothMix against with various state-of-the-art robust training methods for smoothed classifiers on a wide range of image classification benchmarks, including MNIST \cite{dataset/mnist}, CIFAR-10 \cite{dataset/cifar}, and ImageNet \cite{dataset/ilsvrc} datasets.
Overall, the results consistently show that our new adversarial training scheme for smoothed classifiers significantly improves the certified robustness compared to existing methods, \eg one of our CIFAR-10 model could largely outperform an existing state-of-the-art result on the average certified robustness in $\ell_2$-radius $0.720\rightarrow0.737$.
Through an extensive ablation study, we also verify that our method is (a) robust to the choice of hyperparameters, and (b) can effectively {trade-off between the accuracy and robustness} \cite{pmlr-v97-zhang19p} of smoothed classifiers.

Overall, our work suggests that the robustness of a classifier should be set {individually} per sample considering its nearby inputs: we approach this problem by leveraging the relationship between the prediction confidence and robustness of smoothed classifiers. Recently, there have been also some initial attempts to incorporate a \emph{sample-wise} treatment for robustness by allowing input-dependent noise scales in randomized smoothing \cite{alfarra2020data, wang2021pretraintofinetune, chen2021instars}. However, our theoretical analysis shows that such an approach would eventually suffer from the curse of dimensionality (Theorem \ref{thm:dimdep} in Appendix~\ref{ap:adaptive_sigma}), highlighting our approach of focusing on a ``better calibration'' as a promising alternative.

\section{Preliminaries}
\label{s:prelim}
We assume an \textit{i.i.d.}\ dataset $\mathcal{D}=\{(x_i, y_i)\}^n_{i=1} \sim P$, where $x_i \in \mathbb{R}^d$ and $y_i \in \mathcal{Y}:=\{1, \cdots, C\}$, and focus on the problem of correctly classifying a given input $x$ into one of $C$ classes. 
Let $f: \mathbb{R}^d\rightarrow \mathcal{Y}$ be a classifier modeled by $f(x):=\argmax_{c\in\mathcal{Y}}F_c (x)$ with $F: \mathbb{R}^{d}\rightarrow \Delta^{C-1}$, where $\Delta^{C-1}$ denotes the probability simplex in $\mathbb{R}^C$. For example, $F$ can be a neural network followed by a softmax layer. 

In the context of \emph{adversarial robustness}, we require $f$ not only to correctly classify $(x, y) \sim P$, but also to be \emph{locally-constant} around $x$, \ie $f$ should not contain any adversarial examples around $x$.
In this respect, one can measure and attempt to maximize the adversarial robustness of a classifier $f$ by considering the \emph{minimum-distance} of adversarial perturbation \cite{moosavi2016deepfool, carlini2017towards, carlini2019evaluating}, namely:
\begin{equation}
\label{eq:avg_min_dist}
    R(f; x, y) := \min_{f (x')\ne y} \|x' - x\|_2.
\end{equation}

\textbf{Randomized smoothing. }
In cases when $f$ is too complex to control its predictions in practice, \eg if $f$ is a neural network on high-dimensional data, directly solving and maximizing \eqref{eq:avg_min_dist} can be hard.
\emph{Randomized smoothing} \cite{pmlr-v97-cohen19c} instead constructs a new classifier $\hat{f}$ from $f$ that is easier to obtain robustness by transforming the base classifier $f$ with a certain \emph{smoothing measure}, where in this paper we focus on the case of Gaussian distributions $\mathcal{N}(0, \sigma^2 I)$: 
\begin{equation}
\label{eq:smoothing}
    \hat{f}(x) := \argmax_{c\in \mathcal{Y}} \mathbb P_{\delta\sim\mathcal N(0,\sigma^2I)}\left(f(x+\delta)=c\right).
\end{equation}
For a given $(x, y)$, \citet{pmlr-v97-cohen19c} have shown that $R(\hat{f}; x, y)$ can be lower-bounded by the \emph{certified radius} $\underline{R}(\hat{f}, x, y)$, which can be derived from the \emph{confidence} of $\hat{f}$ at $x$, namely we denote it by $p_f(x)$: 
\begin{align}
    \label{eq:cr}
    R(\hat{f}; x, y) \ge \sigma \cdot \Phi^{-1}(p_f(x)) =: \underline{R}(\hat{f}, x, y)~\text{where}~p_f(x):= \mathbb{P}_{\delta\sim\mathcal N(0,\sigma^2I)}(f(x+\delta)=\hat{f}(x)),
\end{align}
provided that $\hat{f}(x) = y$, and otherwise $R(\hat{f}; x, y) := 0$.\footnote{ Here, $\Phi$ denotes the cumulative distribution function of the standard normal distribution.
}
This lower bound is known as tight for the $\ell_2$-minimum distance, \eg the bound is optimal for linear classifiers \cite{pmlr-v97-cohen19c}. 

Although randomized smoothing can be applied for any classifier $f:\mathbb{R}^d \rightarrow \mathcal{Y}$, the robustness of smoothed classifiers can vary depending on $p_f$ {as in} \eqref{eq:cr}, \ie how $f$ performs on a given input under the presence of Gaussian noise. In this sense, to obtain a robust $\hat{f}$, \citet{pmlr-v97-cohen19c} simply propose to train $f$ using Gaussian augmentation by default:
\begin{equation}
\label{eq:gaussian_training}
    \min_{F}\ \mathbb{E}_{\substack{(x, y)\sim P \\ \delta\sim\mathcal{N}(0, \sigma^2 I)}} \left[\mathcal{L}(F(x+\delta), y)\right],
\end{equation}
where $\mathcal{L}$ denotes the standard cross-entropy loss.

\textbf{Adversarial training for smoothed classifiers. }
To obtain $f$ that gives a more robust classifier when smoothed into $\hat{f}$, \citet{nips_salman19} propose \emph{SmoothAdv} that employs adversarial training \cite{madry2018towards} on $\hat{f}$:
\begin{equation}
\label{eq:smoothadv_train}
    \min_{\hat{f}} \max_{\|x' - x\|_2 \le \epsilon}\mathcal{L}(\hat{f}; x', y).
\end{equation} 
Due to the {intractability} of $\hat{f}$, however, it is hard to directly optimize the inner maximization of \eqref{eq:smoothadv_train} via gradient methods. 
To bypass this, SmoothAdv attacks the \emph{soft-smoothed} classifier $\hat{F}:=\mathbb{E}_\delta[F_y(x+\delta)]$ instead. Specifically, SmoothAdv finds an adversarial example via solving the following:
\begin{align}
    \hat{x} = \argmax_{\|x' - x\|_2 \le \epsilon}\mathcal{L}(\hat{F}; x', y)
    \approx \argmax_{\|x' - x\|_2 \le \epsilon}\left(-\log \left(\frac{1}{m} \sum_i F_y(x'+\delta_i)\right)\right),
    \label{eq:smoothadv_train_approx}
\end{align}
using Monte Carlo integration with $m$ samples of $\delta$, namely $\delta_1, \cdots, \delta_m \sim \mathcal{N}(0, \sigma^2 I)$. 

\section{Method}
\label{s:method}
Our goal in this paper is to develop a more suitable form of adversarial training (AT) for smoothed classifiers, taking into account their unique characteristics on adversarial robustness over standard neural networks.
Figure~\ref{fig:concept} illustrates a motivating example: as shown in Figure~\ref{fig:at}, AT typically assumes a fixed-sized ball of radius $\varepsilon$ that each adversarial perturbation must be in, as the goal of the training is to defend the classifier against adversaries under a specific threat model. {However, in a case when AT is applied to a smoothed classifier, {\eg} as done by SmoothAdv, this assumption may be too restrictive, particularly for inputs where the classifier already certifies robustness of radii larger than $\varepsilon$ (\eg Figure~\ref{fig:smoothadv}).
This demands for a new form of AT specially for smoothed classifiers, \eg that allows more distant adversarial examples, despite its fundamental difficulty in the context of standard neural networks {\cite{kang2020testing, pmlr-v119-zhang20z}.}}

In this regard, our proposed training method of \emph{SmoothMix} takes a completely different approach to incorporate adversarial examples during training. More specifically, for a given sample $(x, y) \sim P$, our method finds an adversarial example of $x$ \emph{without} an explicit norm constraint, \ie ``unrestrictively.'' This is because our focus is not to find an input for correcting its label to $y$ (as in the standard AT), but to find an input that is \emph{over-confident} and \emph{semantically off-class}, \ie it is not beneficial to the classifier to label this input to $y$.
Once we have such an example, SmoothMix then labels it as the \emph{uniform confidence}, and considers a \emph{mixup} training \cite{zhang2018mixup} with the original $x$: by linearly interpolating with the uniform confidence, SmoothMix effectively calibrates the over-confident inputs in between, re-balancing the certified radius at the original sample of $x$ at the end.

\subsection{Exploring over-confident adversarial examples in smoothed classifiers} 
\label{ss:overconfident}

Recall that we have a (base) classifier $f$ of the form $f(x)=\argmax_{c\in\mathcal{Y}}F_c (x)$, $\hat{f}$ is its smoothed counterpart, and we aim to improve the robustness of $\hat{f}$ by incorporating adversarial examples in training. 
In this paper, we are particularly interested in adversarial examples of $\hat{f}$ that is found \emph{without} a hard restriction in its perturbation size. More concretely, for a given training sample $(x, y) \sim P$, we find adversarial examples by solving the following optimization:
\begin{equation}
\label{eq:unres_adv}
    \tilde{x} := \argmax_{x'}\left({\mathcal{L}(\hat{f}; x', y) - \beta \cdot \|x' - x\|_2^2}\right),
\end{equation} 
where $\mathcal{L}$ is the cross-entropy loss, and $\beta > 0$ is to ensure that \eqref{eq:unres_adv} cannot be arbitrarily far from $x$. 

As proposed by \citet{nips_salman19} (see Section~\ref{s:prelim}), one can optimize \eqref{eq:unres_adv} by approximating the intractable $\hat{f}$ with the soft-smoothed classifier $\hat{F}:=\mathbb{E}_\delta[F(x+\delta)]$, in a similar manner to \eqref{eq:smoothadv_train_approx}. Based on this approximation, we simply perform a $T$-step gradient ascent from $\tilde{x}^{(0)} := x$ with step size $\alpha > 0$ to solve \eqref{eq:unres_adv} using $m$ samples of $\delta$, namely $\delta_1, \cdots, \delta_m \sim \mathcal{N}(0, \sigma^2 I)$:\footnote{{Here, we note that the $\beta$-term in \eqref{eq:unres_adv} are omitted in \eqref{eq:adv_step}.
In practice, we do not use nor tune $\beta$ in our method mainly for simplicity, as the role of $\beta$ can be replaced by assuming a finite $\alpha\cdot T$, \ie by the \emph{Lagrangian duality}: an unconstrained optimization with $\ell_2$-regularization implicitly defines a hard constraint in its $\ell_2$-norm.}
}
\begin{align}
\label{eq:adv_step}
    \tilde{x}^{(t + 1)} := \tilde{x}^{(t)} + \alpha \cdot \frac{\nabla_x J(\tilde{x}^{(t)})}{\|\nabla_x J(\tilde{x}^{(t)})\|_2}, 
    \text{ where \ } J(x) :=  -\log \left(\frac{1}{m} \sum_i F_y(x+\delta_i)\right).
\end{align}

\begin{figure}
\centering
\begin{minipage}{.65\textwidth}
  \centering
	\hfill
	\subfigure[In-class translation]
	{
	    \includegraphics[width=.45\linewidth]{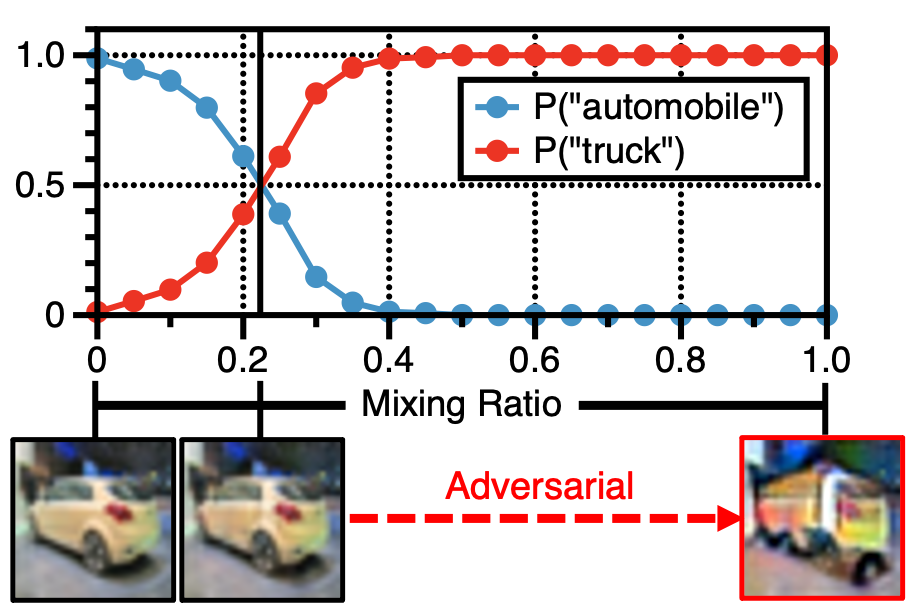}
		\label{fig:conf_in}
	}
	\hfill
	\subfigure[Out-of-class translation]
	{
	    \includegraphics[width=.45\linewidth]{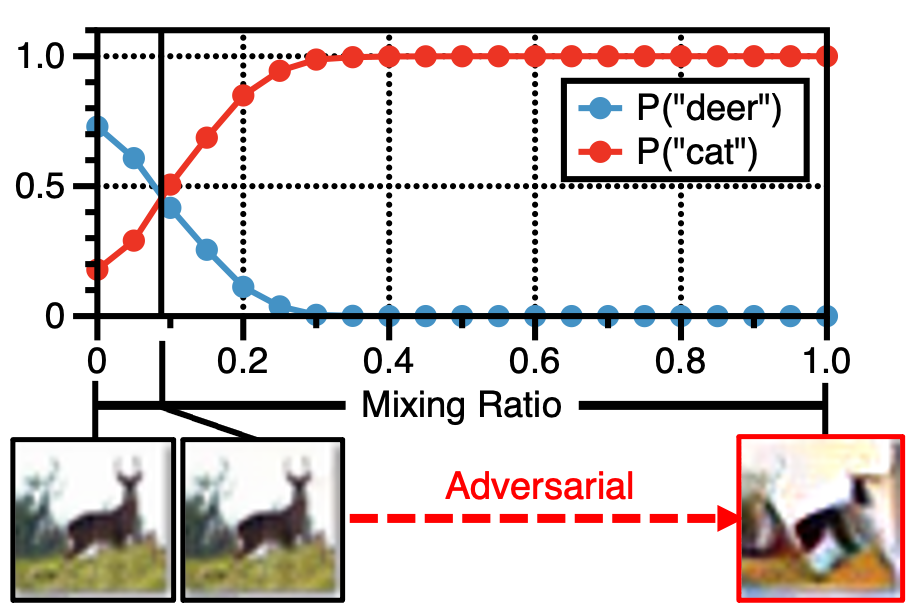}
		\label{fig:conf_out}
	}
	\hfill
	\caption{Illustration of adversarial examples unrestrictively found in CIFAR-10 with a smoothed ResNet-110 ($\sigma=0.25$). The plot demonstrates the change of confidence between two classes as the input is linearly interpolated.}
	\label{fig:mixup_conf}
\end{minipage}
\hfill
\begin{minipage}{.33\textwidth}
  \centering
	\vspace{0.08in}
	\includegraphics[width=0.98\linewidth]{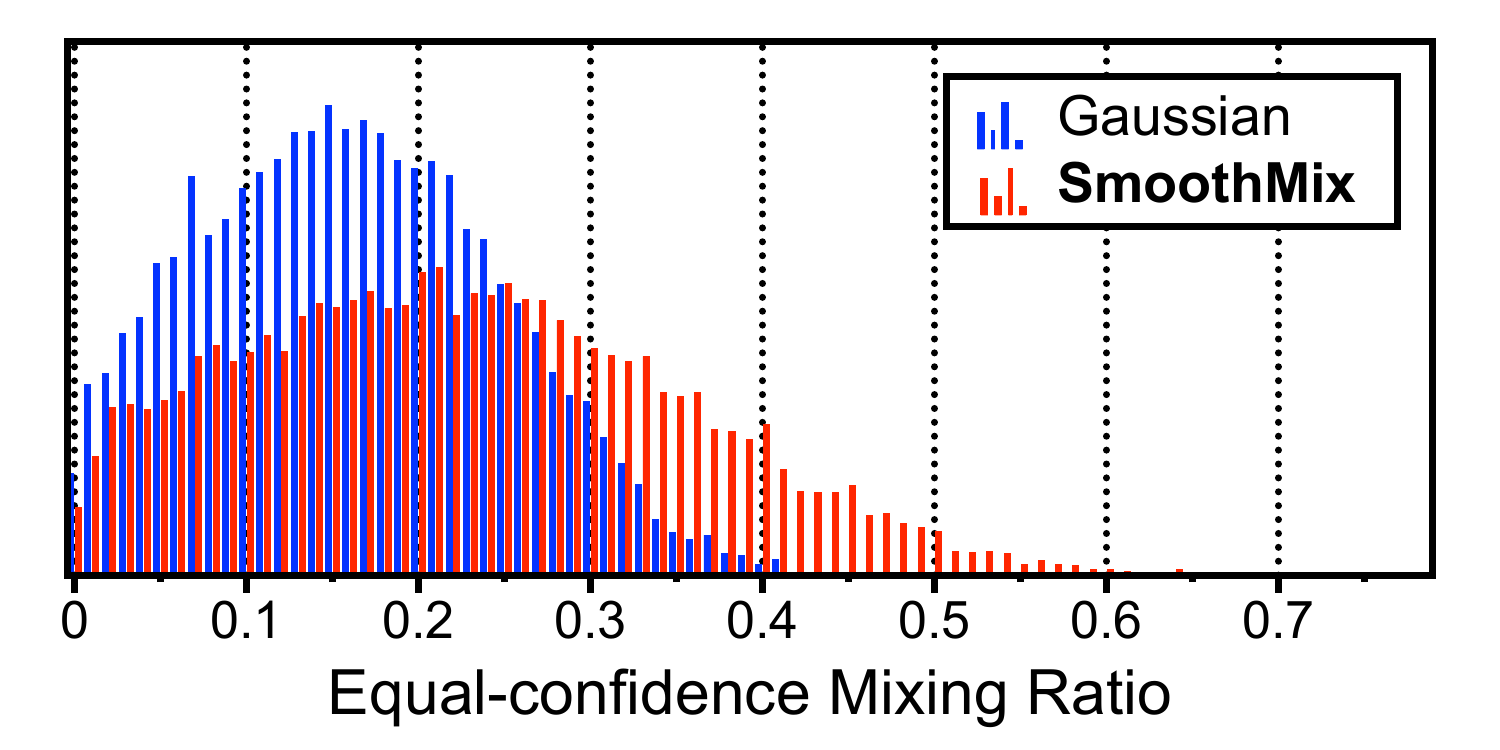}
	\caption{\emph{Equal-confidence mixing ratios} on CIFAR-10, \ie the minimal mixing ratios for changing the correct prediction when each input is linearly interpolated to its adversarial example.}
	\label{fig:mix_lbd}
\end{minipage}
\end{figure}

Figure~\ref{fig:mixup_conf} demonstrates two particular instances of these ``unrestricted'' adversarial examples found from \eqref{eq:adv_step} on $x$, and plots how the confidence of inputs changes as they are linearly interpolated from the clean input to its adversarial counterpart. From this illustration, we make several remarks those would lead to a more direct motivation to our method:
\begin{itemize}
    \item {We observe that adversarial perturbations found via \eqref{eq:adv_step}, \ie from a smoothed classifier, could contain enough amount of semantic changes even in a perceptual sense, in either ways of translating the input to another class (Figure~\ref{fig:conf_in}), or simply removing some relevant information for the current class (Figure~\ref{fig:conf_out}). At least for these cases, therefore, it is reasonable for the classifier to keep its low confidence to the original class. Such a ``perceptually-aligned'' representation is not a unique property of smoothed classifiers, but has been generally observed on adversarially-robust classifiers \cite{santurkar2019cvrobust, engstrom2019adversarial, kaur2019perceptually}: in other words, we leverage the provable robustness of smoothed classifiers during training to reasonably obtain a semantically off-class samples, those will be labeled as the uniform confidence.}
    \item {A major problem we rather highlight here is the tendency of \emph{over-confidence} \cite{pereyra2017regularizing} toward the direction of adversarial perturbation: Figure~\ref{fig:mixup_conf} also presents how the confidence of the given smoothed classifier changes as we linearly interpolate the input from $x$ to $\tilde{x}$. Overall, the adversarially-crafted samples $\tilde{x}$ usually attain significantly higher confidence compared to that of $x$, consequently their \emph{certified radius} \eqref{eq:cr} would be much larger as well. Therefore, considering that $\tilde{x}$ are still nearby $x$, such the over-confidence at $\tilde{x}$ would negatively affect the certified radius of $x$, especially when $\tilde{x}$ does not contain much semantically meaningful information as observed in Figure~\ref{fig:conf_out}.\footnote{We nevertheless remark that such $\tilde x$ is still sufficiently far from $x$ compared to the adversarial examples commonly used in the standard adversarial training (and SmoothAdv), that has a hard $\ell_2$-norm restriction.} \jh{This observation is further supported qualitatively in Table~\ref{tab:confidence}: by comparing the average (a) \emph{true-class} confidence and their (b) \emph{off-class} confidences of a smoothed classifier, those are defined by (a) $\mathbb{E}[\mathbb{P}(f(x+\delta)=y)]$ and (b) $\mathbb{E}[\max_{c\neq y} \mathbb{P}(f(x+\delta)=c)]$, respectively, we confirm that the off-class confidence of $\tilde{x}$ can be abnormally higher than those of clean samples $x$ as we allow more budget on $\varepsilon$.}}
\end{itemize}

\begin{table*}[t]
\centering
\caption{\jh{Comparison of the average (a) true-class confidence, and (b) maximum off-class confidence of smoothed classifiers, for adversarial inputs searched via PGD from CIFAR-10 test samples around $\ell_2$-ball of radius $\varepsilon$. We use ResNet-110 trained on CIFAR-10 with $\sigma=0.5$ for this comparison.}}
\label{tab:confidence}
\vspace{0.03in}
\small
    \begin{tabular}{l|c|ccccc}
    \toprule
    CIFAR-10 (Test set; \%) & Clean & $\varepsilon=1.0$ & $\varepsilon=2.0$ & $\varepsilon=3.0$ & $\varepsilon=4.0$ & $\varepsilon=5.0$  \\ 
    \midrule
    (a) $\mathbb{E}[\mathbb{P}(f(x+\delta)=y)]$ & \textbf{66.4} & 47.1 & 24.3 & 14.2 & 11.3 & 10.7 \\
    (b) $\mathbb{E}[\max_{c\neq y} \mathbb{P}(f(x+\delta)=c)]$ & 24.2 & 37.8 & 59.5 & \textbf{71.8} & \textbf{78.5} & \textbf{82.0} \\
    \bottomrule
\end{tabular}
\end{table*}

\subsection{SmoothMix for confidence-calibrated training of smoothed classifiers}
\label{ss:smoothmix}

{Based on the observations from Section~\ref{ss:overconfident}, we hypothesize that the \emph{miscalibration} of confidences between $x$ and its unrestricted adversarial example $\tilde{x}$ is an important factor that degrades the certified robustness of smoothed classifiers, and propose to penalize the over-confidence by mixing the \emph{uniform} confidence to them. More concretely, we consider the \emph{mixup} \cite{zhang2018mixup} training between $x$ and $\tilde{x}$, \ie by augmenting the given training data with the following pairs:
\begin{align}
\label{eq:mixup}
    x^{\tt mix} := (1 - \lambda) \cdot x + \lambda \cdot \tilde{x}^{(T)}, 
    y^{\tt mix} := (1 - \lambda) \cdot \hat{F}(x) + \lambda \cdot \tfrac{\mathbbm{1}}{C}, \text{ where \ } \lambda \sim \mathcal{U}\left(\left[0, \tfrac{1}{2}\right]\right)  
\end{align}
where $\hat{F}(x)\in\Delta^{C-1}$ is the soft-smoothed prediction of $x$, $\mathcal{U}$ denotes the uniform distribution, $\lambda$ is a random variable that represents the mixing ratio between $(x, \hat{F}(x))$ and $(\tilde{x}^{(T)}, \mathcal{U}(\mathcal{Y}))$, {and $\mathbbm{1}$ denotes the $C$-dimensional vector of ones}. 
Here, we notice that $\lambda$ is sampled only from $[0, \tfrac{1}{2}]$, unlike the standard choice \cite{zhang2018mixup} of $\mathcal{U}([0, 1])$: 
recall from Figure~\ref{fig:conf_in} that $\tilde{x}$ can be often semantically in-class, so that a direct supervision of the uniform confidence on it could harm the classifier. By simply taking only the half part of the mixed samples closer to $x$, we could reasonably avoid these cases while maintaining its effect to prevent the over-confidence issue. The actual loss to minimize for these new data simply follows the cross-entropy loss with Gaussian augmentation, similarly to \eqref{eq:gaussian_training}:}
\begin{equation}
\label{eq:mix_loss}
    L^{\tt mix} := \mathbb{E}_{\delta\sim\mathcal{N}(0, \sigma^2 I)}\left[\mathcal{L}(F(x^{\tt mix}+\delta), y^{\tt mix})\right].
\end{equation}
Recall the over-confidence issue observed in Figure~\ref{fig:mixup_conf} as we follow from $x$ to $\tilde{x}^{(T)}$.
Minimizing $L^{\tt mix}$ \eqref{eq:mix_loss} directly corresponds to calibrating the high confidence at $\tilde{x}^{(T)}$ and the samples in-between, while keeping the original prediction of $\hat{F}(x)$ at $x$. 
\jh{Even in cases that $\tilde{x}^{(T)}$ does not have an over-confidence issue, \ie when its prediction is already close to the uniform confidence, the loss \eqref{eq:mix_loss} would assign a relatively low value for $\tilde{x}^{(T)}$ so that it can act only if there exists an overconfident $\tilde{x}^{(T)}$ nearby $x$.}

\textbf{Incorporating SmoothAdv for free. }
As our method focuses on adversarial examples that are moderately far from the original inputs assuming that the classifier is already locally-smooth, one may still enjoy the effectiveness of SmoothAdv if it could further enforce the local smoothness. We indeed observe that the joint training can be helpful for the robustness of smoothed classifiers, but a na\"ive combination of them could incur too much costs for finding separate adversarial examples for each method. Instead, we found that simply taking $x \leftarrow \tilde{x}^{(1)}$ without modifying our current training, \ie using the \emph{single-step adversarial example} found during \eqref{eq:adv_step} instead of the clean sample, can reasonably bring a similar effect. In this respect, we allow SmoothMix to use $(\tilde{x}^{(1)}, y)$ instead of $(x, y)$ depending on demand of more robustness at expense of decreased clean accuracy. 

\textbf{Overall training. }
Combining the proposed loss with the standard Gaussian training \eqref{eq:gaussian_training} gives the full objective to minimize for our training method. For a given sample $(x, y)\sim P$, and by letting $L^{\tt nat} := \mathbb{E}_\delta\left[\mathcal{L}(F(x+\delta), y)\right]$, the final loss of SmoothMix is given by:
\begin{equation}
\label{eq:total_loss}
    L := L^{\tt nat} + \eta\cdot L^{\tt mix},
\end{equation}
where $\eta > 0$ is a hyperparameter to control the trade-off between accuracy and robustness.
Algorithm~\ref{alg:training} in Appendix~\ref{ap:alg} demonstrates a concrete training procedure of SmoothMix using $m$ samples of $\delta$ for the Monte Carlo approximation.

\section{Experiments}
\label{s:experiments}

We evaluate the effectiveness of our method extensively on MNIST \cite{dataset/mnist}, CIFAR-10 \cite{dataset/cifar}, and ImageNet \cite{dataset/ilsvrc}\footnote{Results on the ImageNet dataset can be found in Appendix~\ref{ap:imagenet}.} classification datasets. 
Overall, the results consistently highlight that our newly proposed training can significantly improve the certified robustness of smoothed classifiers compared to existing robust training methods. We point out the improvements are especially remarkable on the certified accuracy at larger perturbations, at which SmoothMix mainly focus compared to prior arts.
We also conduct an ablation study on the proposed method to convey a detailed analysis on the individual components.
The detailed experimental setups, \eg training details, datasets, and hyperparameters for the baseline methods, are specified in Appendix~\ref{ap:details}.

\textbf{Baseline methods. }
We compare our method with a variety of existing techniques proposed for a robust training of smoothed classifiers, as listed in what follows: (a) Gaussian \cite{pmlr-v97-cohen19c}: standard training with Gaussian augmentation; (b) Stability training \cite{li2019stab}: a cross-entropy regularization between $F(x)$ and $F(x+\delta)$; (c) SmoothAdv \cite{nips_salman19}: adversarial training on smoothed classifier; (d) MACER \cite{Zhai2020MACER}: a regularization that maximizes an approximative form of the certified radius \eqref{eq:cr}; and (e) Consistency \cite{jeong2020consistency}: a KL-divergence based regularization that minimizes the variance of $F(x+\delta)$ across $\delta$. Whenever possible, we use the pre-trained models released by authors for our evaluation to reproduce the baselines. The more detailed training configurations are specified in Appendix~\ref{ap:hyper}. 

\textbf{Evaluation metrics. }
Our evaluation of the robustness for a given smoothed classifier $\hat{f}$ is largely based on the protocol proposed by \citet{pmlr-v97-cohen19c}, similarly to prior works \cite{nips_salman19, Zhai2020MACER, jeong2020consistency}:
more concretely, \citet{pmlr-v97-cohen19c} proposed a practical Monte Carlo based certification procedure, namely \textsc{Certify}, that returns the prediction of $\hat{f}$ and a ``safe'' lower bound of certified radius over the randomness of $n$ samples with probability at least $1-\alpha$, or abstains the certification. 

From \textsc{Certify}, we consider two evaluation metrics: (a) the \emph{approximate certified test accuracy} at various radii: the fraction of the test dataset which \textsc{Certify} classifies correctly with radius larger than $r$ without abstaining, and (b) the \emph{average certified radius} (ACR) \cite{Zhai2020MACER}: the average of certified radii returned by \textsc{Certify} on the test dataset counting only the correctly classified samples, namely $\mathrm{ACR} := \frac{1}{|\mathcal{D}_{\tt test}|}\sum_{(x, y)\in\mathcal{D}_{\tt test}}\mathrm{CR}(f, \sigma, x)\cdot \mathbf{1}_{\hat{f}(x)=y}$,
where $\mathcal{D}_{\tt test}$ is the test dataset, and $\mathrm{CR}$ denotes the certified radius from $\textsc{Certify}(f, \sigma, x)$. Here, the latter metric, ACR, is for a better comparison of robustness under trade-off between accuracy and robustness \cite{tsipras2018robustness, pmlr-v97-zhang19p}: \jh{by its definition, ACR naturally assigns 0 for the incorrectly classified test samples, \ie when $\hat{f}(x)\neq y$, so that a decreased clean accuracy of $\hat{f}$ would negatively affect the value of ACR.}
We use the official PyTorch implementation\footnote{\url{https://github.com/locuslab/smoothing}} of \textsc{Certify}, with $n=100,000$, $n_0=100$ and $\alpha=0.001$, following \cite{pmlr-v97-cohen19c, nips_salman19, jeong2020consistency}. 

\begin{table*}[t]
\centering
\caption{Comparison of approximate certified test accuracy (\%) and ACR on MNIST. All the models are trained and evaluated with the same smoothing factor specified by $\sigma$. Each value except ACR indicates the fraction of test samples which have $\ell_2$ certified radius larger than the threshold specified at the top row. We set our results bold-faced whenever the value improves the Gaussian baseline, and underlined whenever the value improves the best among the considered baselines.}
    \vspace{0.03in}
    \begin{adjustbox}{width=1\linewidth}
    \begin{tabular}{clc|cccccccccccc}
    \toprule
    $\sigma$ &  Models (MNIST) & ACR & 0.00 & 0.25 & 0.50 & 0.75 & 1.00 & 1.25 & 1.50 & 1.75 & 2.00 & 2.25 & 2.50 & 2.75 \\ 
    \midrule
    \multirow{9.5}{*}{0.25} & Gaussian \cite{pmlr-v97-cohen19c} & 0.911 & 99.2 & 98.5 & 96.7 & 93.3 & 0.0 & 0.0 & 0.0 & 0.0 & 0.0 & 0.0 & 0.0 & 0.0 \\
    & Stability training  \cite{li2019stab} & 0.915 & 99.3 & 98.6 & 97.1 & 93.8 & 0.0 & 0.0 & 0.0 & 0.0 & 0.0 & 0.0 & 0.0 & 0.0 \\
    & SmoothAdv \cite{nips_salman19} & {{0.932}} & 99.4 & 99.0 & 98.2 & 96.8 & 0.0 & 0.0 & 0.0 & 0.0 & 0.0 & 0.0 & 0.0 & 0.0  \\
    & MACER  \cite{Zhai2020MACER} & 0.920 & 99.3 & 98.7 & 97.5 & 94.8 & 0.0 & 0.0 & 0.0 & 0.0 & 0.0 & 0.0 & 0.0 & 0.0 \\ 
    & Consistency \cite{jeong2020consistency} & {0.928} & {99.5} & {98.9} & {98.0} & {96.0} & 0.0 & 0.0 & 0.0 & 0.0 & 0.0 & 0.0 & 0.0 & 0.0 \\
    \cmidrule(l){2-2} \cmidrule(l){3-3} \cmidrule(l){4-15}
    & \textbf{SmoothMix ($\eta=1.0$)} & {\textbf{0.931}} & \underline{\textbf{99.5}} & \textbf{98.9} & \underline{\textbf{98.2}} & \textbf{96.4} & 0.0 & 0.0 & 0.0 & 0.0 & 0.0 & 0.0 & 0.0 & 0.0 \\
    & \textbf{+ One-step adversary} & \underline{\textbf{0.933}} & \textbf{99.4} & \underline{\textbf{99.0}} & \underline{\textbf{98.2}} & \underline{\textbf{96.9}} & 0.0 & 0.0 & 0.0 & 0.0 & 0.0 & 0.0 & 0.0 & 0.0 \\
    & \textbf{SmoothMix ($\eta=5.0$)} & {\textbf{0.932}} & \textbf{99.4} & \underline{\textbf{99.0}} & \underline{\textbf{98.2}} & \textbf{96.7} & 0.0 & 0.0 & 0.0 & 0.0 & 0.0 & 0.0 & 0.0 & 0.0 \\
    & \textbf{+ One-step adversary} & \underline{\textbf{0.933}} & \textbf{99.3} & \underline{\textbf{99.0}} & \underline{\textbf{98.2}} & \underline{\textbf{97.0}} & 0.0 & 0.0 & 0.0 & 0.0 & 0.0 & 0.0 & 0.0 & 0.0 \\
    \midrule
    \multirow{9.5}{*}{0.50} & Gaussian \cite{pmlr-v97-cohen19c} & 1.553 & 99.2 & 98.3 & 96.8 & 94.3 & 89.7 & 81.9 & 67.3 & 43.6 & 0.0 & 0.0 & 0.0 & 0.0 \\
    & Stability training  \cite{li2019stab} & 1.570 & 99.2 & 98.5 & 97.1 & 94.8 & 90.7 & 83.2 & 69.2 & 45.4 & 0.0 & 0.0 & 0.0 & 0.0 \\
    & SmoothAdv \cite{nips_salman19} & 1.687 & 99.0 & 98.3 & 97.3 & 95.8 & 93.2 & 88.5 & 81.1 & 67.5 & 0.0 & 0.0 & 0.0 & 0.0 \\
    & MACER  \cite{Zhai2020MACER} & 1.594 & 98.5 & 97.5 & 96.2 & 93.7 & 90.0 & 83.7 & 72.2 & 54.0 & 0.0 & 0.0 & 0.0 & 0.0 \\ 
    & Consistency \cite{jeong2020consistency} & {1.657} & 99.2 & {98.6} & {97.6} & {95.9} & {93.0} & {87.8} & {78.5} & {60.5} & 0.0 & 0.0 & 0.0 & 0.0 \\
    \cmidrule(l){2-2} \cmidrule(l){3-3} \cmidrule(l){4-15}
    & \textbf{SmoothMix ($\eta=1.0$)} & \textbf{{1.678}} & 99.0 & \textbf{98.4} & \textbf{97.4} & \textbf{95.7} & \textbf{93.0} & \textbf{88.1} & \textbf{80.0} & \textbf{65.6} & 0.0 & 0.0 & 0.0 & 0.0 \\
    & \textbf{+ One-step adversary} & \underline{\textbf{1.694}} & 98.8 & 98.1 & \textbf{97.1} & \textbf{95.3} & \textbf{92.7} & \textbf{88.3} & \underline{\textbf{81.7}} & \underline{\textbf{69.5}} & 0.0 & 0.0 & 0.0 & 0.0 \\
    & \textbf{SmoothMix ($\eta=5.0$)} & \underline{\textbf{1.694}} & 98.7 & 98.0 & \textbf{97.0} & \textbf{95.3} & \textbf{92.7} & \underline{\textbf{88.5}} & \underline{\textbf{81.8}} & \underline{\textbf{70.0}} & 0.0 & 0.0 & 0.0 & 0.0 \\
    & \textbf{+ One-step adversary} & \textbf{1.685} & 98.2 & 97.5 & 96.3 & \textbf{94.5} & \textbf{91.3} & \textbf{87.4} & \textbf{81.0} & \underline{\textbf{70.7}} & 0.0 & 0.0 & 0.0 & 0.0 \\
    \midrule
    \multirow{9.5}{*}{1.00} & Gaussian \cite{pmlr-v97-cohen19c} & 1.620 & 96.3 & 94.4 & 91.4 & 86.8 & 79.8 & 70.9 & 59.4 & 46.2 & 32.5 & 19.7 & 10.9 & 5.8 \\
    & Stability training  \cite{li2019stab} &  1.634 & 96.5 & 94.6 & 91.6 & 87.2 & 80.7 & 71.7 & 60.5 & 47.0 & 33.4 & 20.6 & 11.2 & 5.9 \\
    & SmoothAdv \cite{nips_salman19} &  1.779 & 95.8 & 93.9 & 90.6 & 86.5 & 80.8 & 73.7 & 64.6 & 53.9 & 43.3 & 32.8 & 22.2 & 12.1  \\
    & MACER  \cite{Zhai2020MACER} & 1.598 & 91.6 & 88.1 & 83.5 & 77.7 & 71.1 & 63.7 & 55.7 & 46.8 & 38.4 & 29.2 & 20.0 & 11.5  \\
    & Consistency \cite{jeong2020consistency} & {1.740} & 95.0 & 93.0 & 89.7 & 85.4 & 79.7 & 72.7 & 63.6 & 53.0 & 41.7 & 30.8 & 20.3 & 10.7 \\
    \cmidrule(l){2-2} \cmidrule(l){3-3} \cmidrule(l){4-15}
    & \textbf{SmoothMix ($\eta=1.0$)} & \underline{\textbf{1.788}} & 95.5 & 93.5 & 90.5 & 86.2 & \textbf{80.6} & \textbf{73.4} & \textbf{64.3} & \textbf{53.7} & \textbf{43.2} & \underline{\textbf{33.5}} & \underline{\textbf{23.9}} & \underline{\textbf{14.1}} \\
    & \textbf{+ One-step adversary} & \underline{\textbf{1.816}} & 94.7 & 92.4 & 89.2 & 84.6 & 79.4 & \textbf{72.5} & \textbf{64.0} & \underline{\textbf{54.5}} & \underline{\textbf{44.8}} & \underline{\textbf{36.2}} & \underline{\textbf{27.4}} & \underline{\textbf{18.7}} \\
    & \textbf{SmoothMix ($\eta=5.0$)} & \underline{\textbf{1.820}} & 93.7 & 91.6 & 88.1 & 83.5 & 77.9 & 70.9 & \textbf{62.7} & \textbf{53.8} & \underline{\textbf{44.8}} & \underline{\textbf{36.6}} & \underline{\textbf{28.9}} & \underline{\textbf{21.5}} \\
    & \textbf{+ One-step adversary} & \underline{\textbf{1.823}} & 93.3 & 90.9 & 87.5 & 83.0 & 77.5 & 70.6 & \textbf{62.7} & \textbf{53.4} & \underline{\textbf{44.9}} & \underline{\textbf{37.1}} & \underline{\textbf{29.3}} & \underline{\textbf{22.4}} \\
    \bottomrule
\end{tabular}

    \end{adjustbox}
    \label{tab:mnist}
\end{table*}

\begin{figure*}[t]
	\centering
	\subfigure[$\sigma=0.25$]
	{
	    \includegraphics[width=0.31\linewidth]{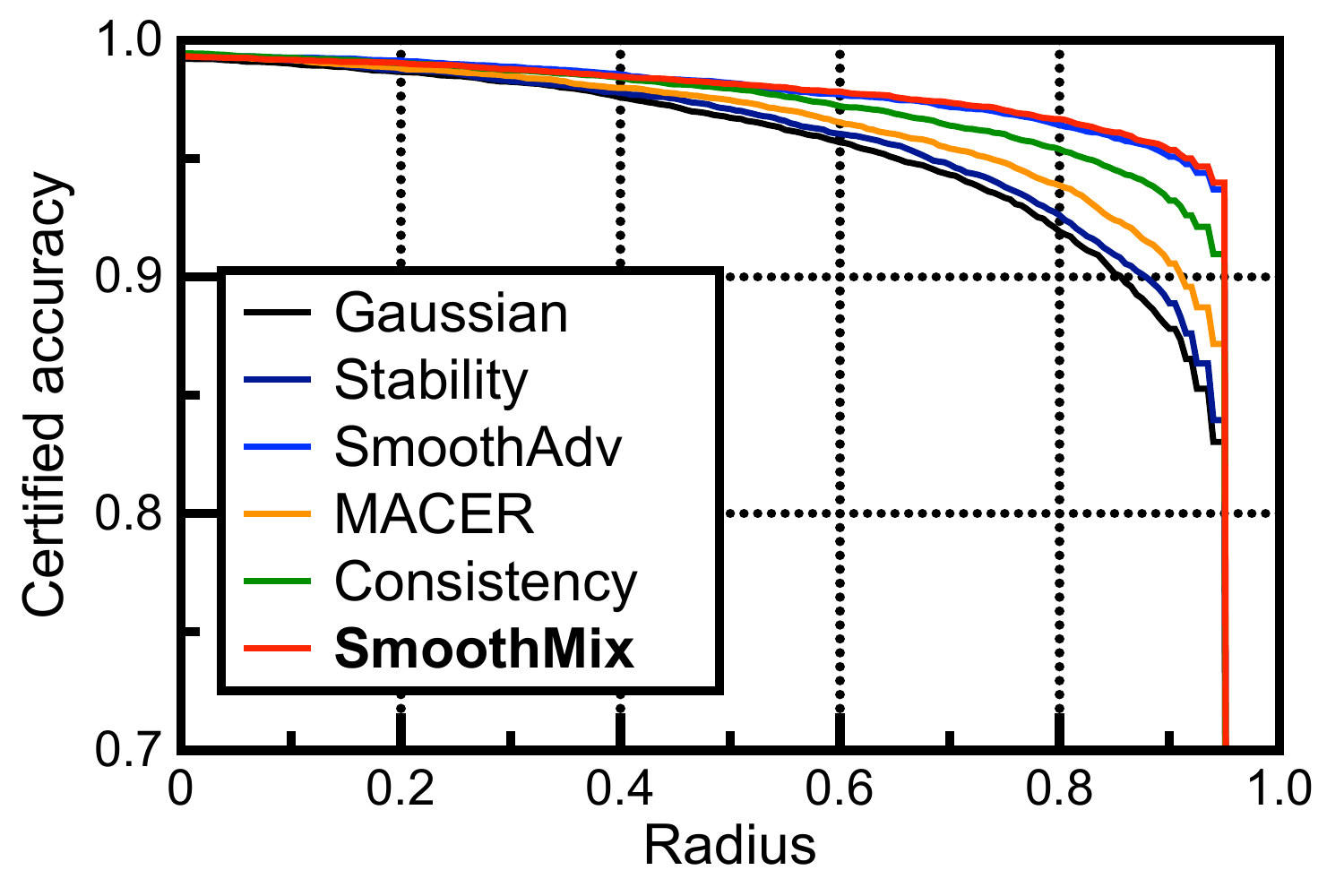}
		\label{fig:mnist_25}
	}
	\subfigure[$\sigma=0.50$]
	{
	    \includegraphics[width=0.31\linewidth]{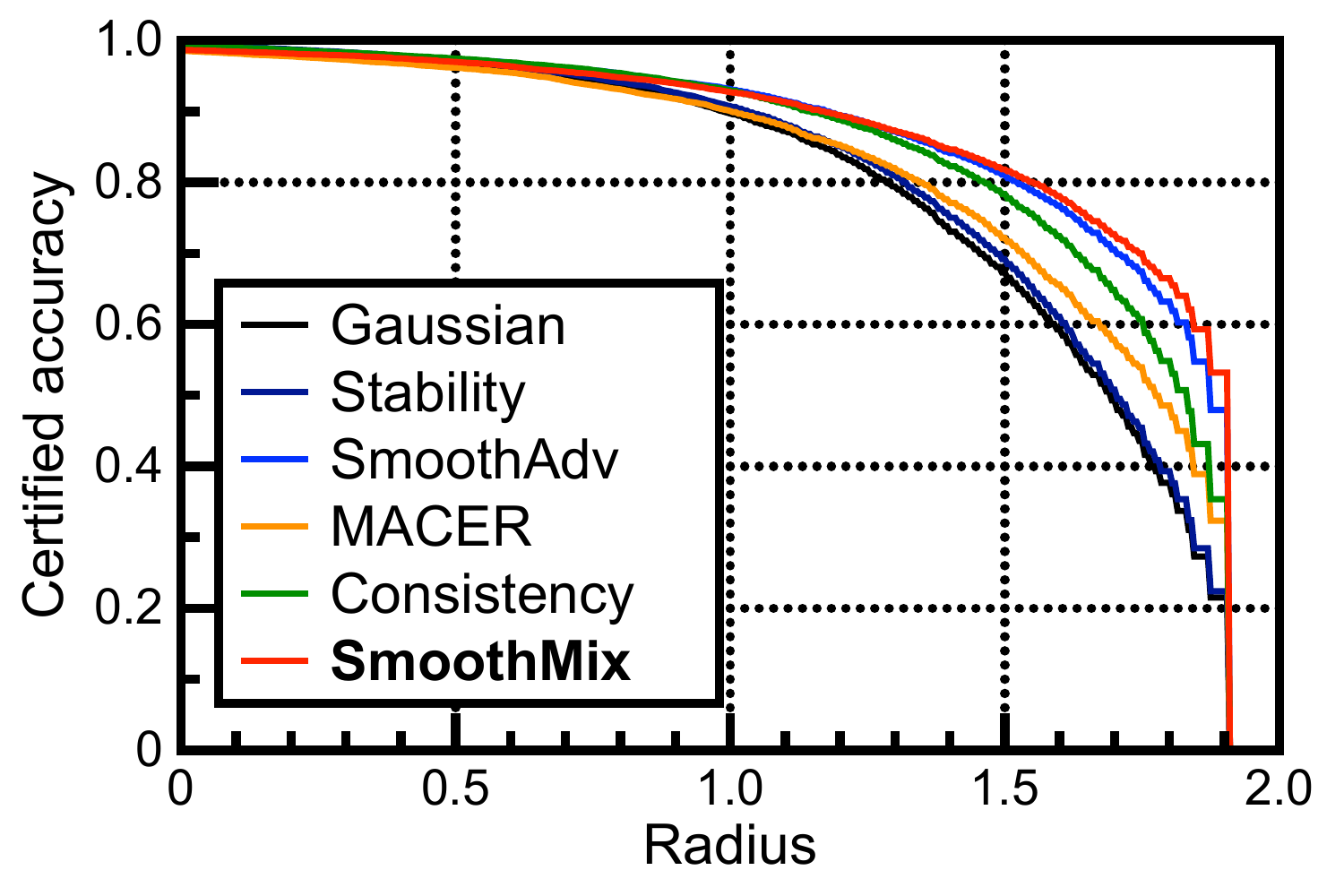}
		\label{fig:mnist_50}
	}
	\subfigure[$\sigma=1.00$]
	{
	    \includegraphics[width=0.31\linewidth]{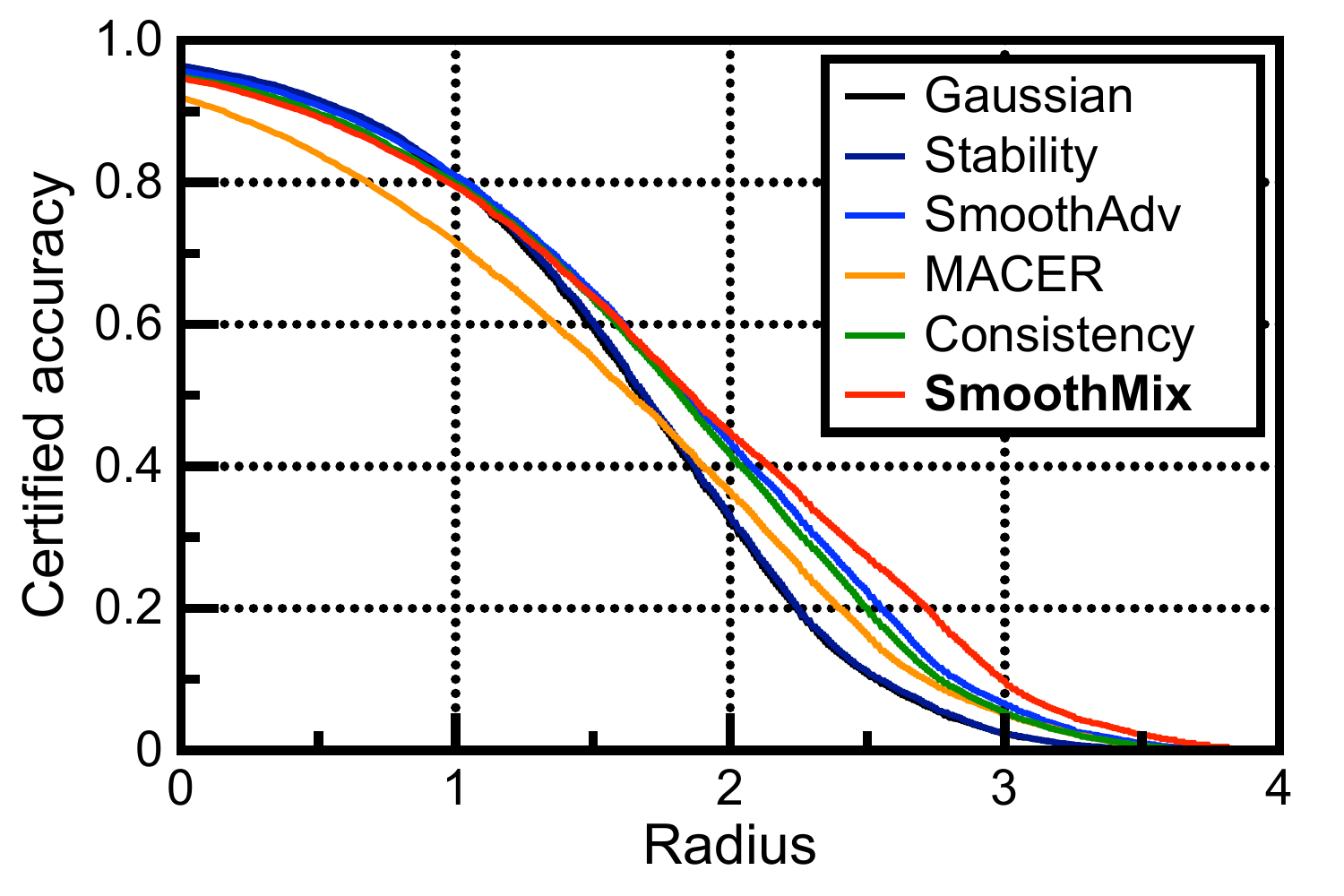}
		\label{fig:mnist_100}
	}
	\caption{Comparison of approximate certified accuracy for various training methods on MNIST. The sharp drop of certified accuracy in each plot is due to that there is a strict upper bound in radius that \textsc{Certify} can output for a given $\sigma$ and $n=100,000$.}
	\label{fig:mnist}
\end{figure*}

\begin{table*}[t]
\centering
\caption{Comparison of approximate certified test accuracy (\%) and ACR on CIFAR-10. We set our results bold-faced whenever the value improves the Gaussian baseline, and underlined whenever the value improves the best among the considered baselines. $^*$~indicates that the results are evaluated from the official pre-trained models released by authors.}
\label{tab:cifar10}
    \vspace{0.03in}
    \small
    \begin{adjustbox}{width=0.82\linewidth}
    \begin{tabular}{clc|cccccccc}
    \toprule
    $\sigma$ &  Models (CIFAR-10) & ACR & 0.00 & 0.25 & 0.50 & 0.75 & 1.00 & 1.25 & 1.50 & 1.75 \\ 
    \midrule
    \multirow{7.5}{*}{0.25}& Gaussian \cite{pmlr-v97-cohen19c} & {0.424} & {76.6} & {61.2} & {42.2} & {25.1} & 0.0 & 0.0 & 0.0 & 0.0  \\
    & Stability training \cite{li2019stab} & 0.421 & 72.3 & 58.0 & 43.3 & 27.3 & 0.0 & 0.0 & 0.0 & 0.0  \\ 
    & SmoothAdv$^*$ \cite{nips_salman19} & {0.544} & {73.4} & {65.6} & {57.0} & {47.5} & 0.0 & 0.0 & 0.0 & 0.0  \\
    & MACER$^*$  \cite{Zhai2020MACER} & {0.531}  & {79.5} & {69.0} & {55.8} & {40.6} & 0.0 & 0.0 & 0.0 & 0.0 \\
    & Consistency \cite{jeong2020consistency} & {{0.552}}  & 75.8 & {67.6} & {58.1} & {46.7} & 0.0 & 0.0 & 0.0 & 0.0  \\
    \cmidrule(l){2-2} \cmidrule(l){3-3} \cmidrule(l){4-11}
    & \textbf{SmoothMix (Ours)} & \underline{\textbf{0.553}} & \textbf{77.1} & \textbf{67.9} & \textbf{57.9} & \textbf{46.7} & 0.0 & 0.0 & 0.0 & 0.0 \\
    & \textbf{+ One-step adversary} & \textbf{0.548} & 74.2 & \textbf{66.1} & \textbf{57.4} & \underline{\textbf{47.7}} & 0.0 & 0.0 & 0.0 & 0.0 \\
    \midrule
    \multirow{7.5}{*}{0.50}& Gaussian \cite{pmlr-v97-cohen19c} & {0.525} & {65.7} & {54.9} & {42.8} & {32.5} & {22.0} & {14.1} & {8.3} & {3.9}  \\
    & Stability training \cite{li2019stab} & {0.521} & 60.6 & 51.5 & 41.4 & 32.5 & 23.9 & 15.3 & 9.6 & 5.0 \\ 
    & SmoothAdv$^*$ \cite{nips_salman19}  & 0.684 & 65.3 & 57.8 & 49.9 & 41.7 & 33.7 & 26.0 & 19.5 & 12.9 \\  
    & MACER$^*$ \cite{Zhai2020MACER} & {0.691} & {64.2} & {57.5} & {49.9} & {42.3} & {34.8} & {27.6} & {20.2} & {12.6}  \\
    & Consistency \cite{jeong2020consistency} & {{0.720}}  & 64.3 & {57.5} & {50.6} & {43.2} & {36.2} & {29.5} & {22.8} & {16.1} \\
    \cmidrule(l){2-2} \cmidrule(l){3-3} \cmidrule(l){4-11}
    & \textbf{SmoothMix (Ours)} & \textbf{0.715} & 65.0 & \textbf{56.7} & \textbf{49.2} & \textbf{41.2} & \textbf{34.5} & \underline{\textbf{29.6}} & \underline{\textbf{23.5}} & \underline{\textbf{18.1}} \\
    & \textbf{+ One-step adversary} & \underline{\textbf{0.737}} & 61.8 & \textbf{55.9} & \textbf{49.5} & \underline{\textbf{43.3}} & \underline{\textbf{37.2}} & \underline{\textbf{31.7}} & \underline{\textbf{25.7}} & \underline{\textbf{19.8}} \\
    \bottomrule
\end{tabular}
    \end{adjustbox}
\end{table*}

\subsection{Results on MNIST}
\label{ss:result_mnist}

For MNIST \cite{dataset/mnist} experiments, we report the approximate certified accuracy and ACR of smoothed classifiers obtained from LeNet \cite{dataset/mnist} with different training methods, including SmoothMix, using the full MNIST test dataset. 
We consider three different models as varying the noise level $\sigma\in\{0.25, 0.5, 1.0\}$. During inference, we apply randomized smoothing with the same $\sigma$ used in the training. 
When SmoothMix is used, we consider a fixed hyperparameter value for $\alpha=1.0$ and $m=4$, the step size and the number of noise samples. We empirically observe that it is beneficial to set $\alpha\cdot T$ to be proportional to $\sigma$, the noise level, as there exist different upper bounds on the certified radius statistical achievable in practice depending on $\sigma$: in this respect, we set $T=2, 4, 8$ for the models with $\sigma=0.25, 0.5, 1.0$, respectively.
We apply the same $m=4$ for SmoothAdv, \ie for adversarial training, as well, and $T=10$ with an $\ell_2$-constraint of radius $\varepsilon=1.0$. 
\jh{Here, notice that SmoothMix and SmoothAdv use the same number of hyperparameters: more specifically, although SmoothMix introduces $\alpha$ compared to SmoothAdv, the step size in \eqref{eq:adv_step}, it instead does not use the hyperparameter $\varepsilon$ of SmoothAdv, \ie the maximum norm of adversarial perturbations.}

The results are presented in Table~\ref{tab:mnist} and Figure~\ref{fig:mnist}. Overall, we observe that our proposed SmoothMix loss \eqref{eq:mix_loss} added to the Gaussian training dramatically improves the certified test accuracy from ``Gaussian''. By considering the one-step adversary (Section~\ref{ss:smoothmix}) in training, we could further improve the robust accuracy, significantly improving ACRs compared to the previous state-of-the-art training methods: \eg our method could improve ACRs with $\sigma=1.0$ from $1.779 \rightarrow 1.823$. This shows that improvements from SmoothMix can be orthogonal to those from SmoothAdv. It is also remarkable that even without the one-step adversarial example, one could further improve the certified robustness by simply increasing the relative strength $\eta$ of the SmoothMix loss, \eg by $1.0 \rightarrow 5.0$ as presented in Table~\ref{tab:mnist}: \eg ``SmoothMix'' with $\eta=5.0$ still outperforms ``SmoothAdv'' by $1.779 \rightarrow 1.820$ at $\sigma=1.0$. Finally, we note that our models could substantially improve the robustness at larger perturbations with less degradation in the clean accuracy, \eg compared to ``MACER'' or ``Consistency'': considering that they are also regularization based approaches that allow to control the robustness via controlling their regularization strength, the results show that our form of loss could better compensate the trade-off between accuracy and robustness.

\subsection{Results on CIFAR-10}
\label{ss:result_cifar10}

For CIFAR-10 \cite{dataset/cifar} experiments, we report the approximate certified accuracy and ACR of smoothed classifiers from ResNet-110 \cite{he2016deep} using the full CIFAR-10 test dataset. Again, we consider three different models as varying the noise level $\sigma\in\{0.25, 0.5, 1.0\}$,\footnote{Due to the space limitation, we defer the CIFAR-10 results with $\sigma=1.0$ to Appendix~\ref{ap:add_cifar10}, considering that the scenario can be less practical compared to the others: \eg the clean accuracy in this setup is $<50\%$ in most cases, even for the Gaussian baseline \cite{pmlr-v97-cohen19c}.} and apply the same $\sigma$ for inference as well. 
When SmoothMix is used, we consider a fixed hyperparameter value for $T=4$ and $m=2$, the number of steps and the number of noise samples, respectively. We also fix $\eta=5.0$ throughout the experiments, as also used for MNIST (see Table~\ref{tab:mnist}). Again, we make sure that $\alpha \cdot T$ to be proportional to $\sigma$, so that we set $\alpha=0.5, 1.0, 2.0$ for the models with $\sigma=0.25, 0.5, 1.0$, respectively.
For SmoothAdv, we report the performance evaluated from the pre-trained models released by the authors\footnote{\url{https://github.com/Hadisalman/smoothing-adversarial}} for a fixed configuration of $T=10, \varepsilon=1.0$, and $m=8$. 

\begin{figure}
\centering
\begin{minipage}{.65\textwidth}
  \centering
	\hfill
	\subfigure[$\sigma=0.25$]
	{
	    \includegraphics[width=0.46\linewidth]{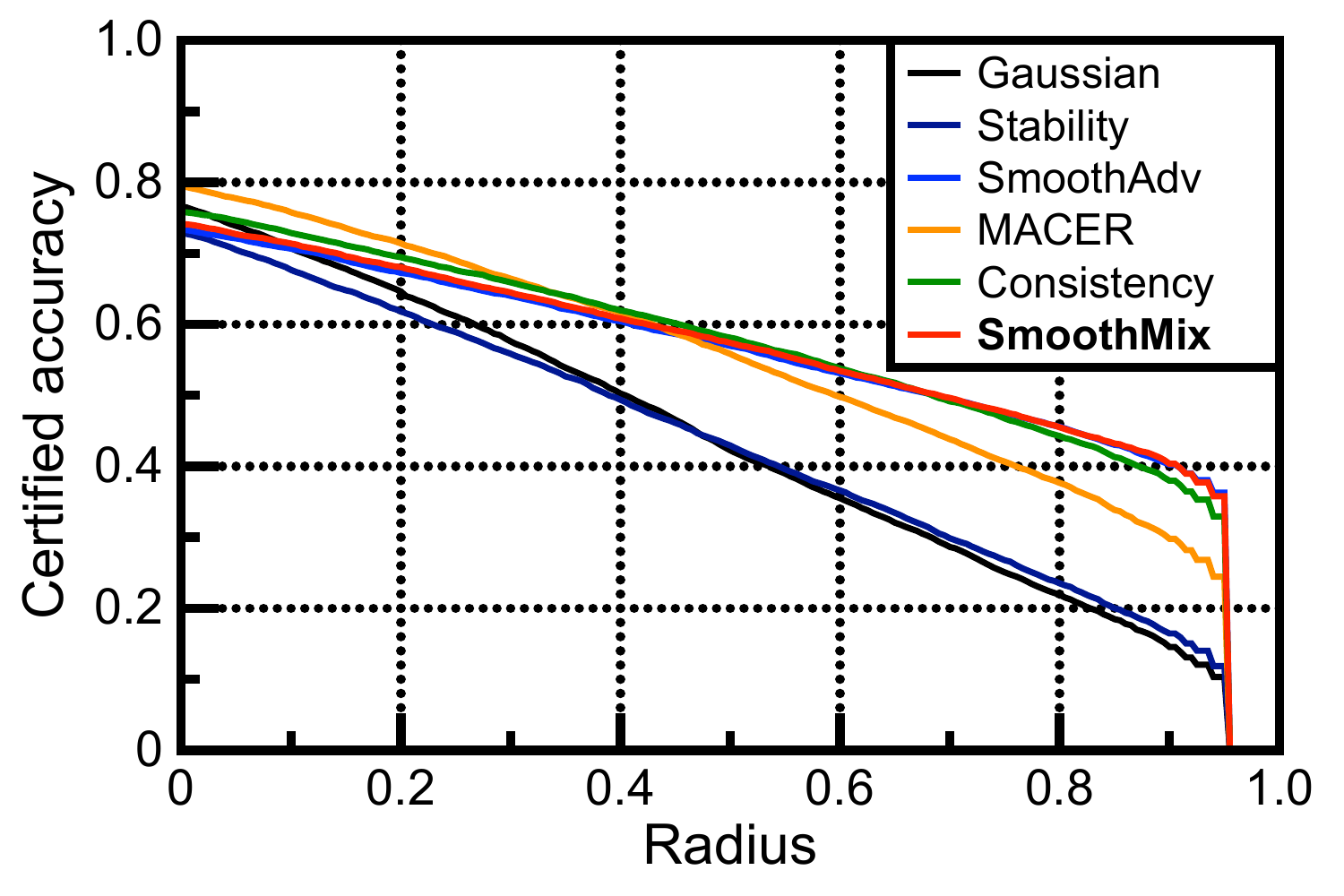}
		\label{fig:cifar10_25}
	}
	\subfigure[$\sigma=0.50$]
	{
	    \includegraphics[width=0.46\linewidth]{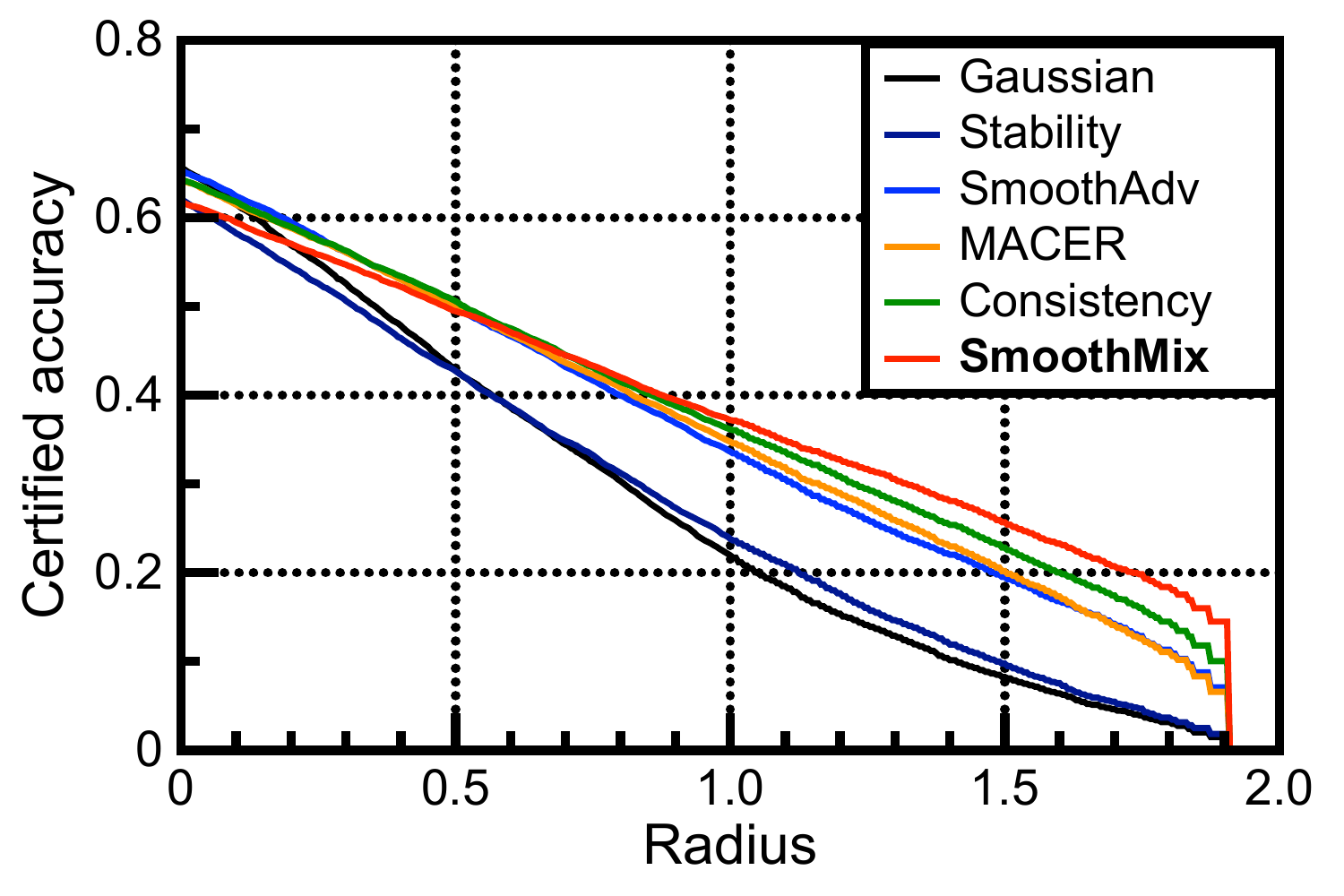}
		\label{fig:cifar10_50}
	}
	\hfill
	\caption{Comparison of approximate certified accuracy for various training methods on CIFAR-10. The sharp drop of certified accuracy in each plot is due to that there is a strict upper bound in radius that \textsc{Certify} can output for a given $\sigma$ and $n=100,000$.}
	\label{fig:cifar10}
\end{minipage}
\hfill
\begin{minipage}{.33\textwidth}
  \centering
	\vspace{0.05in}
	\includegraphics[width=0.92\linewidth]{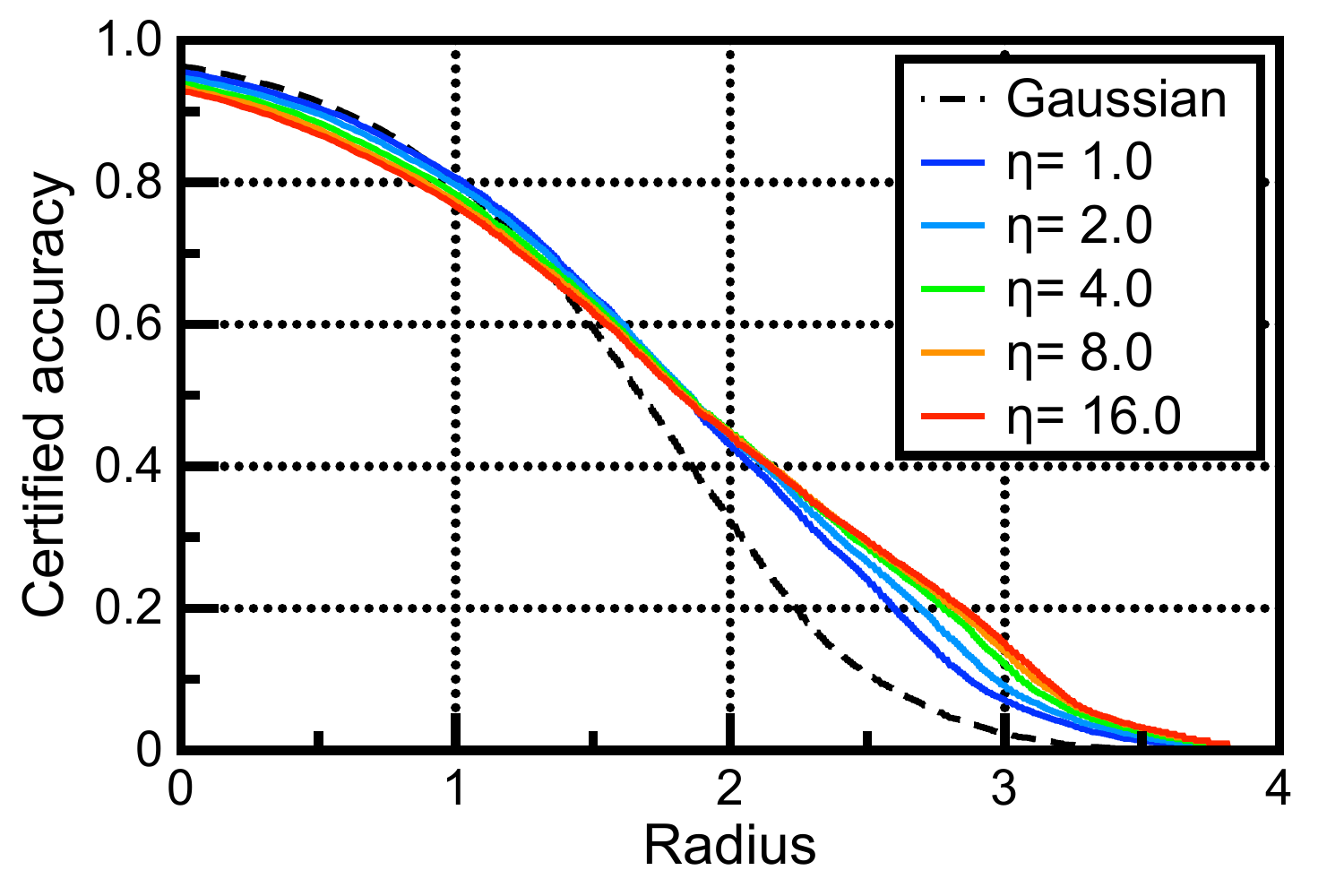}
	\vspace{0.05in}
	\caption{Comparison of approximate certified test accuracy of SmoothMix for varying $\eta$. ``Gaussian'' indicates the baseline training with Gaussian augmentation.}
	\label{fig:eta}
\end{minipage}
\vspace{-0.05in}
\end{figure}

The results are summarized in Table~\ref{tab:cifar10} and Figure~\ref{fig:cifar10}. Again, we still observe that our method generally exhibits better trade-offs between accuracy and certified robustness compared to other baselines: \eg at $\sigma=0.5$, ``SmoothMix'' could improve the previous best result from ``Consistency'' by a significant margin of $0.720 \rightarrow 0.737$. 
Without the single-step adversary, ``SmoothMix'' can effectively preserve the clean accuracy while also improving ACR, \eg at $\sigma=0.25$, ``SmoothMix'' could even improve the clean accuracy of ``Gaussian'': although ``MACER'' could improve the clean accuray as well, one could see that their improvements in robust accuracy are relatively limited.
\jh{It is also notable that the \emph{certified test accuracy} we report in Table~\ref{tab:cifar10} can sometimes complement ACR: although ``Consistecny'' achieves a competitive ACR with ``SmoothMix'' at $\sigma=0.25$, one can still confirm the superiority of ``SmoothMix'' by comparing the certified accuracy $r=0.0$ (\ie the clean accuracy), namely 75.8\% vs.~77.1\%, given that they both achieve similar certified accuracy at $r=0.75$ (\ie the robust accuracy).  This is because a bare increase in the clean accuracy (\ie $\hat{f}$ correctly classifies more test samples but with $\mathrm{CR}$'s closer to 0) often contributes less to the increase in ACR.}

\subsection{Ablation study}
\label{ss:ablation}

We also conduct an ablation study to investigate the individual effects of the hyperparameters in our method.
Unless otherwise noted, we perform experiments on MNIST with $\sigma=1.0$. 
All the detailed results from this ablation study are reported in Appendix~\ref{ap:ablation}.

\textbf{Equal-confidence mixing ratios.}
Recall from Figure~\ref{fig:mixup_conf} that we are motivated by the problem of \emph{miscalibration} in smoothed classifiers between clean and its adversarial example. To see how much the proposed SmoothMix could alleviate this issue, we compare the distributions of the minimal mixing ratios that changes its prediction of a given classifier on the CIFAR-10 test samples, namely the \emph{equal-confidence mixing ratios}, before and after training with SmoothMix. 
\jh{We find the adversarial examples separately from two pre-trained ResNet-110 based (smoothed) classifiers, each trained by Gaussian training and SmoothMix, respectively, assuming $\sigma=0.25$. We optimize each adversarial example assuming only a quite loose norm-bound of $\varepsilon=8$ to allow more update steps, \ie via 50-step PGD for both classifiers.}
Figure~\ref{fig:mix_lbd} shows the result, and it indeed confirms SmoothMix has an effect of improving calibration between clean and adversarial examples.  

\textbf{Effect of $\eta$.}
By design, SmoothMix controls the trade-off between accuracy and robustness by adjusting $\eta$, the relative strength of $L^{\tt mix}$ over $L^{\tt nat}$ \eqref{eq:total_loss}.
Here, we further examine the effect of $\eta$ by comparing the certified robustness on varying $\eta\in\{1, 2, 4, 8, 16\}$:
the results in Figure~\ref{fig:eta} show that increasing $\eta$ consistently improves the certified robustness of the classifier, which confirms $L^{\tt mix}$, the mixup loss, as an effective term to trade-off the robustness against $L^{\tt nat}$ for accuracy. 

\begin{figure*}[t]
	\centering
	\hfill
	\subfigure[Effect of $\alpha$ and $T$]
	{
	    \includegraphics[width=0.31\linewidth]{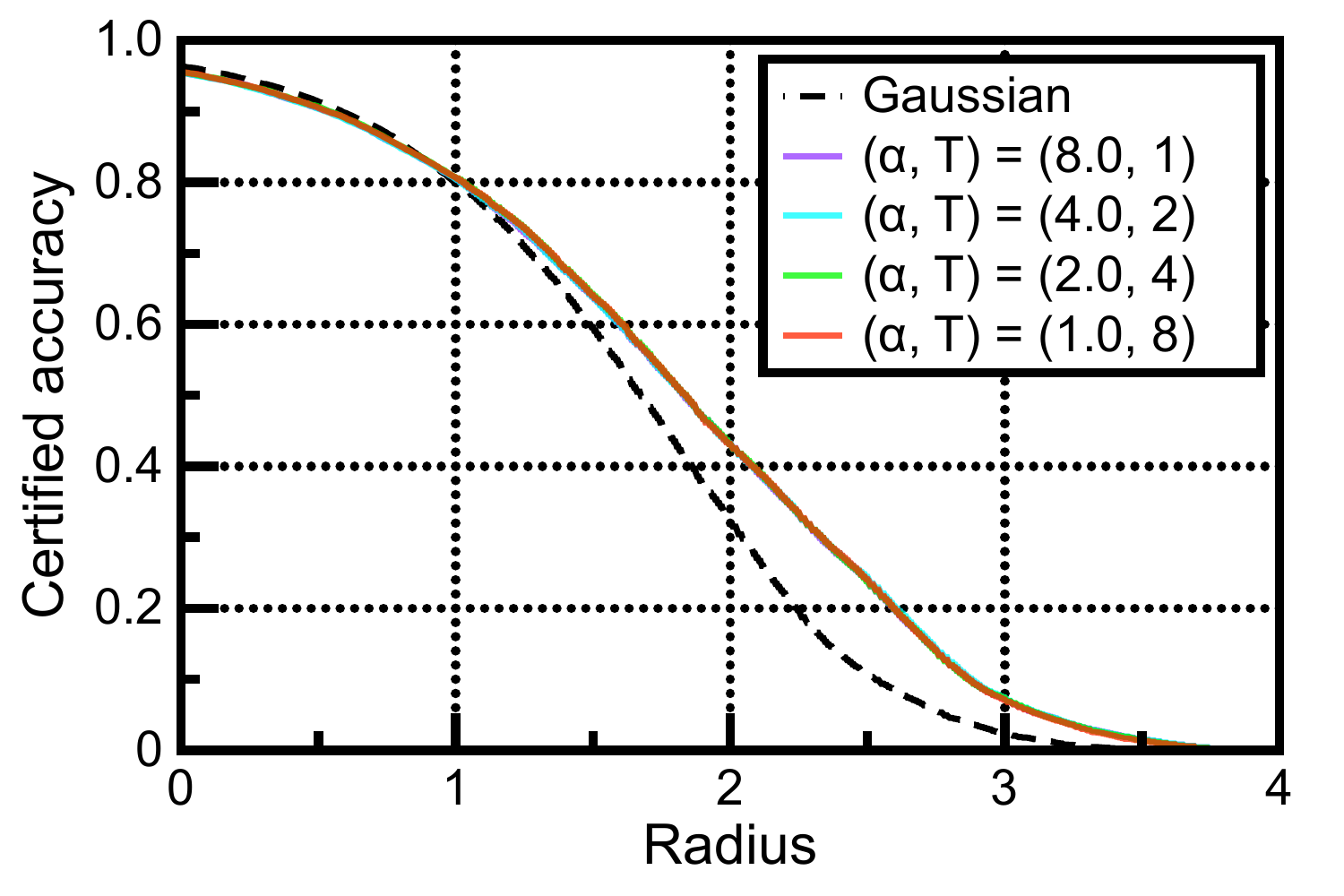}
		\label{fig:alpha_steps}
	}
	\hfill
    \subfigure[Hard restriction of $\varepsilon$ on attacks]
	{
		\includegraphics[width=0.31\linewidth]{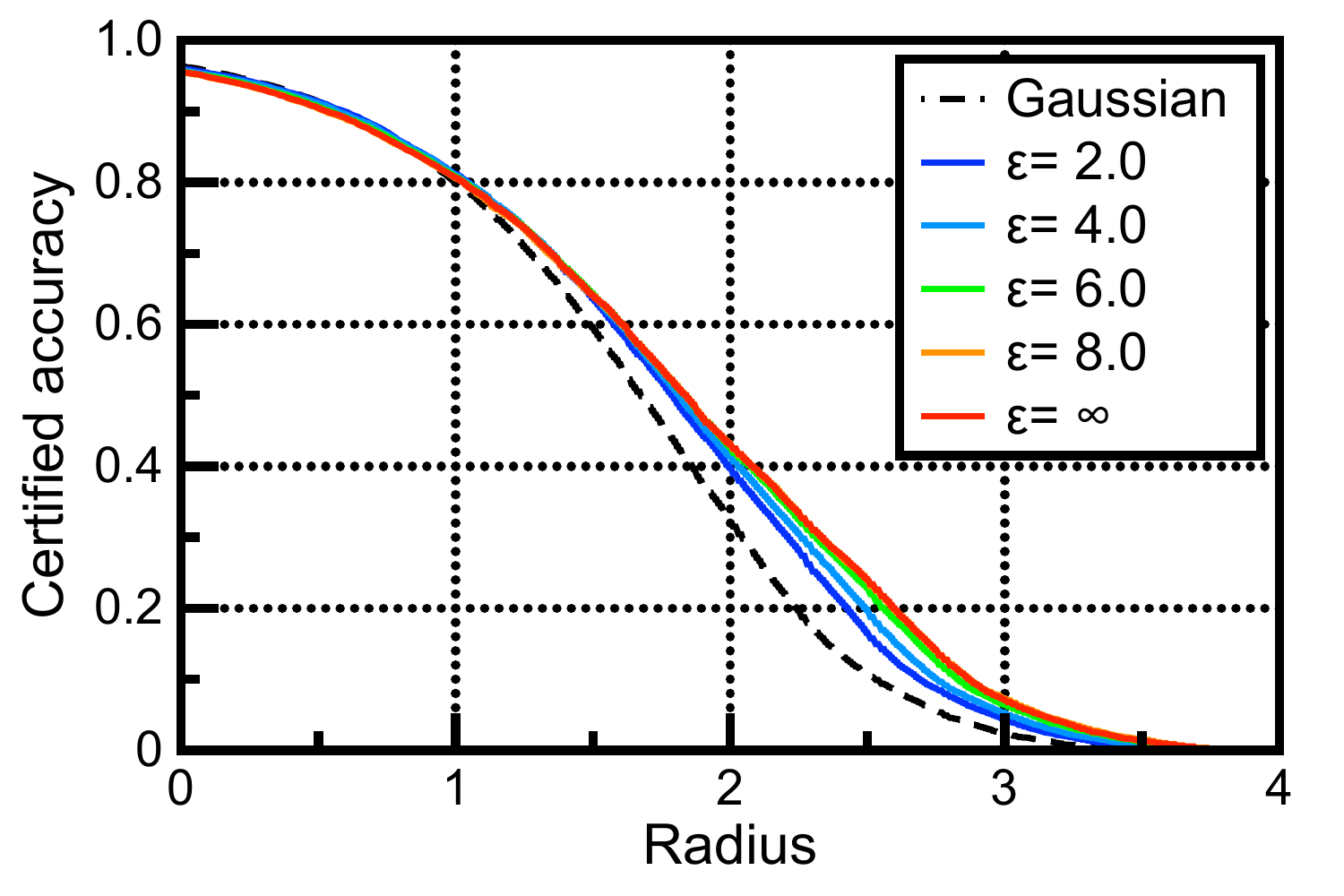}
		\label{fig:maxnorm}
	}
	\hfill
	\subfigure[Effect of $m$]
	{
	    \includegraphics[width=0.31\linewidth]{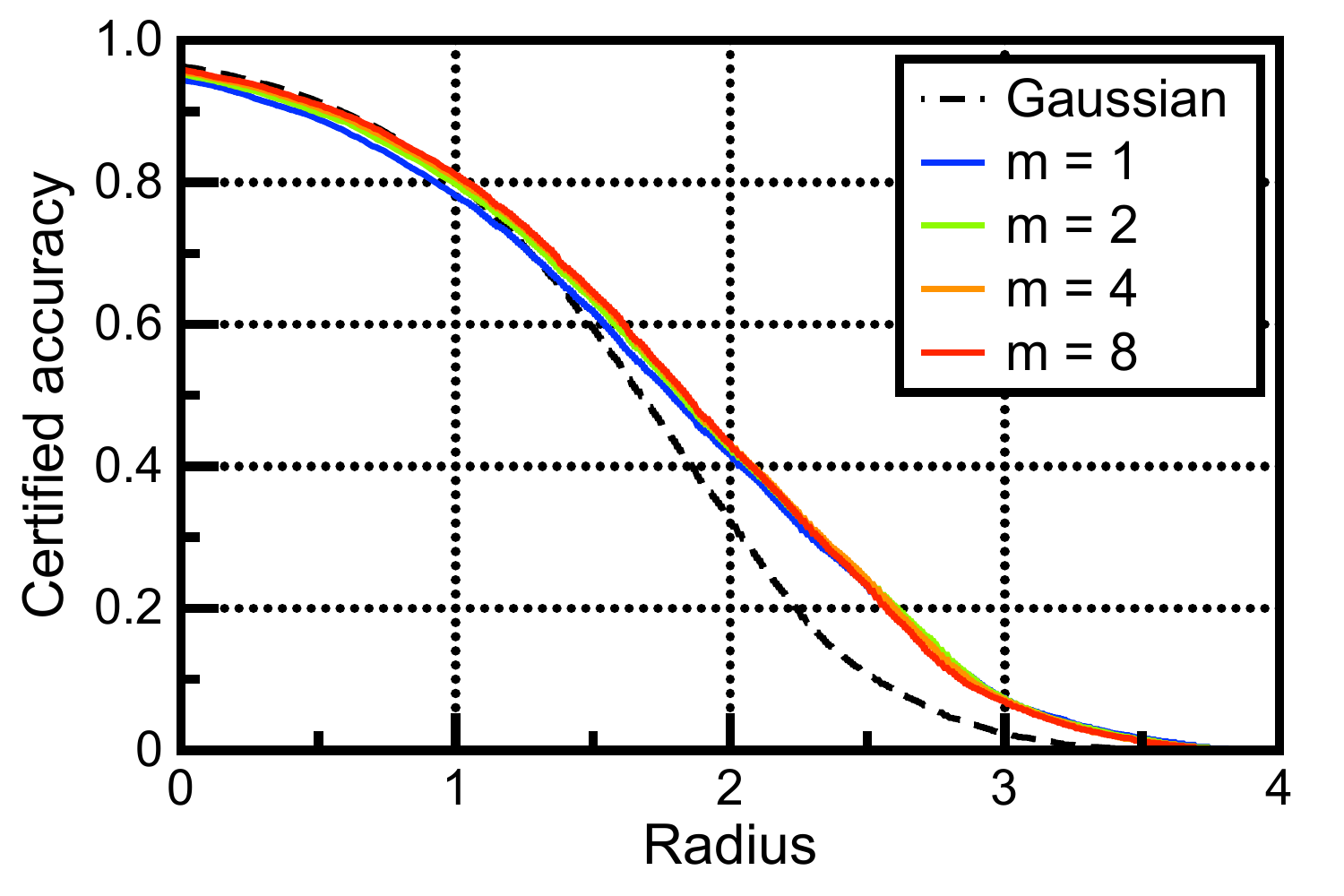}
		\label{fig:noise_vecs}
	}
	\hfill
	\caption{Comparison of approximate certified test accuracy of SmoothMix and its ablations. ``Gaussian'' indicates the baseline training with Gaussian augmentation.}
	\label{fig:ablation}
\end{figure*}

\textbf{Trade-off between $\alpha$ and $T$.}
In practice, SmoothMix can trade-off between the step size $\alpha$ and the number of steps $T$ to compensate between a more accurate optimization of \eqref{eq:unres_adv} and its computational cost, while maintaining the effective range of the perturbation by $\alpha \cdot T$. Figure~\ref{fig:alpha_steps} explores this trade-off, by comparing models trained with different combinations of $(\alpha, T)$ under control of $\alpha \cdot T = 8.0$.
Interestingly, the results indicate that the choice of $\alpha$ and $T$ does not significantly affect the final performance as long as $\alpha \cdot T$ is constant: all the considered combinations achieve similar robustness, with only a slight degradation in ACR even at $(\alpha, T)=(8.0, 1)$ (see Table~\ref{tab:ab_aT} in Appendix~\ref{ap:ablation}). This suggests that (a) finding adversarial examples in a smoothed classifier can be simpler than one might expect, and (b) one can effectively reduce the training cost of SmoothMix using small $T$ in practice. 

\textbf{Hard restriction on adversarial attacks.}
One of key features of SmoothMix is at its \emph{unrestricted} search of adversarial examples. Here, we examine the case when there is a hard restriction on each search, namely in $\ell_2$-radius of $\varepsilon\in\{2, 4, 6, 8\}$. The results presented in Figure~\ref{fig:maxnorm} along with the Gaussian baseline (``Gaussian'') and the original unrestricted setup (``$\varepsilon=\infty$'') show that SmoothMix indeed works best when there is no such restrictions, although these ablations still reasonably improve the Gaussian baseline, \ie calibrating with adversarial examples outside the $\varepsilon$-ball can indeed help to improve the certified robustness in our training scheme.

\textbf{Effect of $m$.}
Figure~\ref{fig:noise_vecs} (and Table~\ref{tab:ab_m} in Appendix~\ref{ap:ablation}) investigates the effect of using different $m\in\{1, 2, 4, 8\}$, the number of noise samples to approximate the prediction of smoothed classifier: the larger $m$, the better approximation of smoothed classifier, which would be beneficial for both natural loss and SmoothMix loss \eqref{eq:total_loss}.
Overall, we observe that SmoothMix can still improve ACR from ``Gaussian'' even with $m=1$, but with a moderate degradation in the clean accuracy: as $m$ is one of the crucial factors related to the total training cost in practice, one is recommended to use smaller $m$, \eg $m=2$ or $4$, considering its little effect to the final ACR.

\section{Discussion and conclusion}
\label{s:conclusion}

We observe that adversarial training with an \emph{unrestricted} adversary can be feasible and even more promising (compared to the \emph{restricted} ones) when it comes with smoothed classifiers, by showing their effectiveness to improve the certified adversarial robustness with a novel mixup-based training. 
We address the brittleness of deep neural networks through the lens of smoothed classifiers, which could give us a simpler view on them. We believe our research could be a useful step toward understanding what essentially constitutes adversarial examples in deep neural networks. 

\textbf{Broader impact. } 
Adversarial robustness in deep learning is arguably an essential requirement for \emph{AI~safety} \cite{amodei2016concrete}, with much impact on various security-concerned systems: \eg medical diagnosis \cite{caruana2015intelligible}, speech recognition \cite{qin2019speech} and autonomous driving \cite{yurtsever2020survey}.
Thanks to their certifiable guarantees, we believe a practical success of systems based on \emph{randomized smoothing} would be fatal for those who maliciously attempt to break down the system via adversarial attacks.
Nevertheless, one should also recognize that current techniques for robustness in deep learning, including randomized smoothing as well, indeed have a clear gap to be practically used in real-world, \eg defending against challenging unrestricted attacks \cite{xiao2018spatially, bhattad2020Unrestricted}, which should be further investigated in the future research.  
Consequently, it is particularly important for the defense techniques not to be misused in practical systems, 
as a failure of such systems may lead practitioners to have a biased, false sense of security.

\textbf{Limitation. } 
While our ultimate goal is to find optimal smoothed classifiers in terms of the accuracy and robustness trade-off, SmoothMix should not be considered as the final solution for the problem; our method is a promising proof-of-concept showing the close relationship between randomized smoothing and \emph{confidence-calibrated} classifiers \cite{pmlr-v70-guo17a, lee2018training}. Although our focus in this paper is currently limited only to the over-confidence issue, we believe there are still many rooms to be explored in future for another such connection, \eg could the recent developments in the literature of uncertainty estimation of deep neural networks \cite{hendrycks2019self, tack2020csi} help to improve the robustness of smoothed classifiers.

\begin{ack}

This work was conducted by Center for Applied Research in Artificial Intelligence (CARAI) grant funded by Defense Acquisition Program Administration (DAPA) and Agency for Defense Development (ADD) (UD190031RD). 
The authors would like to thank Jaeho Lee for helpful discussions.

% Use unnumbered first level headings for the acknowledgments. All acknowledgments
% More information about this disclosure can be found at: \url{https://neurips.cc/Conferences/2021/PaperInformation/FundingDisclosure}.

% Do {\bf not} include this section in the anonymized submission, only in the final paper. You can use the \texttt{ack} environment provided in the style file to autmoatically hide this section in the anonymized submission.
\end{ack}

\bibliographystyle{plainnat}
\bibliography{references}

\begin{thebibliography}{70}
\providecommand{\natexlab}[1]{#1}
\providecommand{\url}[1]{\texttt{#1}}
\expandafter\ifx\csname urlstyle\endcsname\relax
  \providecommand{\doi}[1]{doi: #1}\else
  \providecommand{\doi}{doi: \begingroup \urlstyle{rm}\Url}\fi

\bibitem[Alfarra et~al.(2020)Alfarra, Bibi, Torr, and Ghanem]{alfarra2020data}
Motasem Alfarra, Adel Bibi, Philip H.~S. Torr, and Bernard Ghanem.
\newblock Data dependent randomized smoothing, 2020.

\bibitem[Amodei et~al.(2016)Amodei, Olah, Steinhardt, Christiano, Schulman, and
  Man{\'e}]{amodei2016concrete}
Dario Amodei, Chris Olah, Jacob Steinhardt, Paul Christiano, John Schulman, and
  Dan Man{\'e}.
\newblock Concrete problems in {AI} safety.
\newblock \emph{arXiv preprint arXiv:1606.06565}, 2016.

\bibitem[Athalye et~al.(2018)Athalye, Carlini, and Wagner]{pmlr-v80-athalye18a}
Anish Athalye, Nicholas Carlini, and David Wagner.
\newblock Obfuscated gradients give a false sense of security: Circumventing
  defenses to adversarial examples.
\newblock In \emph{Proceedings of the 35th International Conference on Machine
  Learning}, volume~80 of \emph{Proceedings of Machine Learning Research},
  pages 274--283, Stockholmsmässan, Stockholm Sweden, 10--15 Jul 2018. PMLR.
\newblock URL \url{http://proceedings.mlr.press/v80/athalye18a.html}.

\bibitem[Balunovic and Vechev(2020)]{Balunovic2020Adversarial}
Mislav Balunovic and Martin Vechev.
\newblock Adversarial training and provable defenses: Bridging the gap.
\newblock In \emph{International Conference on Learning Representations}, 2020.
\newblock URL \url{https://openreview.net/forum?id=SJxSDxrKDr}.

\bibitem[Bhattad et~al.(2020)Bhattad, Chong, Liang, Li, and
  Forsyth]{bhattad2020Unrestricted}
Anand Bhattad, Min~Jin Chong, Kaizhao Liang, Bo~Li, and D.~A. Forsyth.
\newblock Unrestricted adversarial examples via semantic manipulation.
\newblock In \emph{International Conference on Learning Representations}, 2020.
\newblock URL \url{https://openreview.net/forum?id=Sye_OgHFwH}.

\bibitem[Carlini and Wagner(2017{\natexlab{a}})]{carlini2017adversarial}
Nicholas Carlini and David Wagner.
\newblock Adversarial examples are not easily detected: Bypassing ten detection
  methods.
\newblock In \emph{Proceedings of the 10th ACM Workshop on Artificial
  Intelligence and Security}, pages 3--14, 2017{\natexlab{a}}.

\bibitem[Carlini and Wagner(2017{\natexlab{b}})]{carlini2017towards}
Nicholas Carlini and David Wagner.
\newblock Towards evaluating the robustness of neural networks.
\newblock In \emph{IEEE Symposium on Security and Privacy}, pages 39--57. IEEE,
  2017{\natexlab{b}}.

\bibitem[Carlini et~al.(2019)Carlini, Athalye, Papernot, Brendel, Rauber,
  Tsipras, Goodfellow, and Madry]{carlini2019evaluating}
Nicholas Carlini, Anish Athalye, Nicolas Papernot, Wieland Brendel, Jonas
  Rauber, Dimitris Tsipras, Ian Goodfellow, and Aleksander Madry.
\newblock On evaluating adversarial robustness.
\newblock \emph{arXiv preprint arXiv:1902.06705}, 2019.

\bibitem[Caruana et~al.(2015)Caruana, Lou, Gehrke, Koch, Sturm, and
  Elhadad]{caruana2015intelligible}
Rich Caruana, Yin Lou, Johannes Gehrke, Paul Koch, Marc Sturm, and Noemie
  Elhadad.
\newblock Intelligible models for healthcare: Predicting pneumonia risk and
  hospital 30-day readmission.
\newblock In \emph{Proceedings of the 21th ACM SIGKDD International Conference
  on Knowledge Discovery and Data Mining}, pages 1721--1730, 2015.

\bibitem[Chen et~al.(2021)Chen, Kong, Yu, Luque, Goldstein, and
  Huang]{chen2021instars}
Chen Chen, Kezhi Kong, Peihong Yu, Juan Luque, Tom Goldstein, and Furong Huang.
\newblock {Insta-RS}: Instance-wise randomized smoothing for improved
  robustness and accuracy, 2021.

\bibitem[Cohen et~al.(2019)Cohen, Rosenfeld, and Kolter]{pmlr-v97-cohen19c}
Jeremy Cohen, Elan Rosenfeld, and Zico Kolter.
\newblock Certified adversarial robustness via randomized smoothing.
\newblock In Kamalika Chaudhuri and Ruslan Salakhutdinov, editors,
  \emph{Proceedings of the 36th International Conference on Machine Learning},
  volume~97 of \emph{Proceedings of Machine Learning Research}, pages
  1310--1320, Long Beach, California, USA, 09--15 Jun 2019. PMLR.
\newblock URL \url{http://proceedings.mlr.press/v97/cohen19c.html}.

\bibitem[Croce and Hein(2020)]{Croce2020Provable}
Francesco Croce and Matthias Hein.
\newblock Provable robustness against all adversarial $l_p$-perturbations for
  $p\geq 1$.
\newblock In \emph{International Conference on Learning Representations}, 2020.
\newblock URL \url{https://openreview.net/forum?id=rklk_ySYPB}.

\bibitem[Croce et~al.(2019)Croce, Andriushchenko, and Hein]{pmlr-v89-croce19a}
Francesco Croce, Maksym Andriushchenko, and Matthias Hein.
\newblock Provable robustness of {ReLU} networks via maximization of linear
  regions.
\newblock In Kamalika Chaudhuri and Masashi Sugiyama, editors,
  \emph{Proceedings of Machine Learning Research}, volume~89 of
  \emph{Proceedings of Machine Learning Research}, pages 2057--2066. PMLR,
  16--18 Apr 2019.
\newblock URL \url{http://proceedings.mlr.press/v89/croce19a.html}.

\bibitem[Dvijotham et~al.(2020)Dvijotham, Hayes, Balle, Kolter, Qin, Gyorgy,
  Xiao, Gowal, and Kohli]{dvijotham2020a}
Krishnamurthy~(Dj) Dvijotham, Jamie Hayes, Borja Balle, Zico Kolter, Chongli
  Qin, Andras Gyorgy, Kai Xiao, Sven Gowal, and Pushmeet Kohli.
\newblock A framework for robustness certification of smoothed classifiers
  using f-divergences.
\newblock In \emph{International Conference on Learning Representations}, 2020.
\newblock URL \url{https://openreview.net/forum?id=SJlKrkSFPH}.

\bibitem[Engstrom et~al.(2019)Engstrom, Ilyas, Santurkar, Tsipras, Tran, and
  Madry]{engstrom2019adversarial}
Logan Engstrom, Andrew Ilyas, Shibani Santurkar, Dimitris Tsipras, Brandon
  Tran, and Aleksander Madry.
\newblock Adversarial robustness as a prior for learned representations.
\newblock \emph{arXiv preprint arXiv:1906.00945}, 2019.

\bibitem[Gehr et~al.(2018)Gehr, Mirman, Drachsler-Cohen, Tsankov, Chaudhuri,
  and Vechev]{gehr2018ai2}
Timon Gehr, Matthew Mirman, Dana Drachsler-Cohen, Petar Tsankov, Swarat
  Chaudhuri, and Martin Vechev.
\newblock Ai2: Safety and robustness certification of neural networks with
  abstract interpretation.
\newblock In \emph{2018 IEEE Symposium on Security and Privacy}, pages 3--18.
  IEEE, 2018.

\bibitem[Goodfellow et~al.(2015)Goodfellow, Shlens, and
  Szegedy]{goodfellow2014explaining}
Ian~J Goodfellow, Jonathon Shlens, and Christian Szegedy.
\newblock Explaining and harnessing adversarial examples.
\newblock In \emph{International Conference on Learning Representations}, 2015.

\bibitem[Gowal et~al.(2019)Gowal, Dvijotham, Stanforth, Bunel, Qin, Uesato,
  Arandjelovic, Mann, and Kohli]{gowal2019scalable}
Sven Gowal, Krishnamurthy~Dj Dvijotham, Robert Stanforth, Rudy Bunel, Chongli
  Qin, Jonathan Uesato, Relja Arandjelovic, Timothy Mann, and Pushmeet Kohli.
\newblock Scalable verified training for provably robust image classification.
\newblock In \emph{Proceedings of the IEEE/CVF International Conference on
  Computer Vision}, pages 4842--4851, 2019.

\bibitem[Guo et~al.(2017)Guo, Pleiss, Sun, and Weinberger]{pmlr-v70-guo17a}
Chuan Guo, Geoff Pleiss, Yu~Sun, and Kilian~Q. Weinberger.
\newblock On calibration of modern neural networks.
\newblock In Doina Precup and Yee~Whye Teh, editors, \emph{Proceedings of the
  34th International Conference on Machine Learning}, volume~70 of
  \emph{Proceedings of Machine Learning Research}, pages 1321--1330,
  International Convention Centre, Sydney, Australia, 06--11 Aug 2017. PMLR.
\newblock URL \url{http://proceedings.mlr.press/v70/guo17a.html}.

\bibitem[He et~al.(2016)He, Zhang, Ren, and Sun]{he2016deep}
Kaiming He, Xiangyu Zhang, Shaoqing Ren, and Jian Sun.
\newblock Deep residual learning for image recognition.
\newblock In \emph{Proceedings of the IEEE Conference on Computer Vision and
  Pattern Recognition}, pages 770--778, 2016.

\bibitem[Hendrycks and Gimpel(2017)]{hendrycks2016baseline}
Dan Hendrycks and Kevin Gimpel.
\newblock A baseline for detecting misclassified and out-of-distribution
  examples in neural networks.
\newblock In \emph{International Conference on Learning Representations}, 2017.
\newblock URL \url{https://openreview.net/forum?id=Hkg4TI9xl}.

\bibitem[Hendrycks et~al.(2019)Hendrycks, Mazeika, Kadavath, and
  Song]{hendrycks2019self}
Dan Hendrycks, Mantas Mazeika, Saurav Kadavath, and Dawn Song.
\newblock Using self-supervised learning can improve model robustness and
  uncertainty.
\newblock In \emph{Advances in Neural Information Processing Systems},
  volume~32. Curran Associates, Inc., 2019.
\newblock URL
  \url{https://proceedings.neurips.cc/paper/2019/file/a2b15837edac15df90721968986f7f8e-Paper.pdf}.

\bibitem[Jeong and Shin(2020)]{jeong2020consistency}
Jongheon Jeong and Jinwoo Shin.
\newblock Consistency regularization for certified robustness of smoothed
  classifiers.
\newblock In \emph{Advances in Neural Information Processing Systems}, 2020.

\bibitem[Jiang et~al.(2018)Jiang, Kim, Guan, and Gupta]{NEURIPS2018_7180cffd}
Heinrich Jiang, Been Kim, Melody Guan, and Maya Gupta.
\newblock To trust or not to trust a classifier.
\newblock In S.~Bengio, H.~Wallach, H.~Larochelle, K.~Grauman, N.~Cesa-Bianchi,
  and R.~Garnett, editors, \emph{Advances in Neural Information Processing
  Systems}, volume~31, pages 5541--5552. Curran Associates, Inc., 2018.
\newblock URL
  \url{https://proceedings.neurips.cc/paper/2018/file/7180cffd6a8e829dacfc2a31b3f72ece-Paper.pdf}.

\bibitem[Kang et~al.(2020)Kang, Sun, Hendrycks, Brown, and
  Steinhardt]{kang2020testing}
Daniel Kang, Yi~Sun, Dan Hendrycks, Tom Brown, and Jacob Steinhardt.
\newblock Testing robustness against unforeseen adversaries, 2020.

\bibitem[Kaur et~al.(2019)Kaur, Cohen, and Lipton]{kaur2019perceptually}
Simran Kaur, Jeremy Cohen, and Zachary~C Lipton.
\newblock Are perceptually-aligned gradients a general property of robust
  classifiers?
\newblock \emph{arXiv preprint arXiv:1910.08640}, 2019.

\bibitem[Kim et~al.(2020)Kim, Choo, and Song]{pmlr-v119-kim20b}
Jang-Hyun Kim, Wonho Choo, and Hyun~Oh Song.
\newblock Puzzle {Mix}: Exploiting saliency and local statistics for optimal
  mixup.
\newblock In Hal~Daumé III and Aarti Singh, editors, \emph{Proceedings of the
  37th International Conference on Machine Learning}, volume 119 of
  \emph{Proceedings of Machine Learning Research}, pages 5275--5285. PMLR,
  13--18 Jul 2020.
\newblock URL \url{http://proceedings.mlr.press/v119/kim20b.html}.

\bibitem[Kim et~al.(2021)Kim, Choo, Jeong, and Song]{kim2021comixup}
JangHyun Kim, Wonho Choo, Hosan Jeong, and Hyun~Oh Song.
\newblock {Co-Mixup}: Saliency guided joint mixup with supermodular diversity.
\newblock In \emph{International Conference on Learning Representations}, 2021.
\newblock URL \url{https://openreview.net/forum?id=gvxJzw8kW4b}.

\bibitem[Krizhevsky(2009)]{dataset/cifar}
Alex Krizhevsky.
\newblock Learning multiple layers of features from tiny images.
\newblock Technical report, Department of Computer Science, University of
  Toronto, 2009.

\bibitem[Kumar et~al.(2019)Kumar, Liang, and Ma]{NEURIPS2019_f8c0c968}
Ananya Kumar, Percy~S Liang, and Tengyu Ma.
\newblock Verified uncertainty calibration.
\newblock In \emph{Advances in Neural Information Processing Systems},
  volume~32, pages 3792--3803. Curran Associates, Inc., 2019.
\newblock URL
  \url{https://proceedings.neurips.cc/paper/2019/file/f8c0c968632845cd133308b1a494967f-Paper.pdf}.

\bibitem[Lamb et~al.(2019)Lamb, Verma, Kannala, and
  Bengio]{lamb2019interpolated}
Alex Lamb, Vikas Verma, Juho Kannala, and Yoshua Bengio.
\newblock Interpolated adversarial training: Achieving robust neural networks
  without sacrificing too much accuracy.
\newblock In \emph{Proceedings of the 12th ACM Workshop on Artificial
  Intelligence and Security}, pages 95--103, 2019.

\bibitem[{Le{C}un} et~al.(1998){Le{C}un}, {Bottou}, {Bengio}, and
  {Haffner}]{dataset/mnist}
Y.~{Le{C}un}, L.~{Bottou}, Y.~{Bengio}, and P.~{Haffner}.
\newblock Gradient-based learning applied to document recognition.
\newblock \emph{Proceedings of the IEEE}, 86\penalty0 (11):\penalty0
  2278--2324, Nov 1998.
\newblock ISSN 1558-2256.
\newblock \doi{10.1109/5.726791}.

\bibitem[Lecuyer et~al.(2019)Lecuyer, Atlidakis, Geambasu, Hsu, and
  Jana]{lecuyer2019certified}
Mathias Lecuyer, Vaggelis Atlidakis, Roxana Geambasu, Daniel Hsu, and Suman
  Jana.
\newblock Certified robustness to adversarial examples with differential
  privacy.
\newblock In \emph{2019 IEEE Symposium on Security and Privacy (SP)}, pages
  656--672. IEEE, 2019.

\bibitem[Lee et~al.(2019)Lee, Yuan, Chang, and Jaakkola]{lee2019tight}
Guang-He Lee, Yang Yuan, Shiyu Chang, and Tommi Jaakkola.
\newblock Tight certificates of adversarial robustness for randomly smoothed
  classifiers.
\newblock In \emph{Advances in Neural Information Processing Systems},
  volume~32, pages 4910--4921. Curran Associates, Inc., 2019.

\bibitem[Lee et~al.(2017)Lee, Hwang, Park, and Shin]{pmlr-v70-lee17b}
Kimin Lee, Changho Hwang, KyoungSoo Park, and Jinwoo Shin.
\newblock Confident multiple choice learning.
\newblock In Doina Precup and Yee~Whye Teh, editors, \emph{Proceedings of the
  34th International Conference on Machine Learning}, volume~70 of
  \emph{Proceedings of Machine Learning Research}, pages 2014--2023,
  International Convention Centre, Sydney, Australia, 06--11 Aug 2017. PMLR.
\newblock URL \url{http://proceedings.mlr.press/v70/lee17b.html}.

\bibitem[Lee et~al.(2018)Lee, Lee, Lee, and Shin]{lee2018training}
Kimin Lee, Honglak Lee, Kibok Lee, and Jinwoo Shin.
\newblock Training confidence-calibrated classifiers for detecting
  out-of-distribution samples.
\newblock In \emph{International Conference on Learning Representations}, 2018.
\newblock URL \url{https://openreview.net/forum?id=ryiAv2xAZ}.

\bibitem[Lee et~al.(2020)Lee, Lee, and Yoon]{lee2020adversarial}
Saehyung Lee, Hyungyu Lee, and Sungroh Yoon.
\newblock Adversarial vertex mixup: Toward better adversarially robust
  generalization.
\newblock In \emph{Proceedings of the IEEE/CVF Conference on Computer Vision
  and Pattern Recognition}, pages 272--281, 2020.

\bibitem[Li et~al.(2019)Li, Chen, Wang, and Carin]{li2019stab}
Bai Li, Changyou Chen, Wenlin Wang, and Lawrence Carin.
\newblock Certified adversarial robustness with additive noise.
\newblock In \emph{Advances in Neural Information Processing Systems 32}, pages
  9464--9474. Curran Associates, Inc., 2019.

\bibitem[Li et~al.(2020)Li, Qi, Xie, and Li]{li2020sokcertified}
Linyi Li, Xiangyu Qi, Tao Xie, and Bo~Li.
\newblock {SoK}: Certified robustness for deep neural networks.
\newblock \emph{arXiv preprint arXiv:2009.04131}, 2020.

\bibitem[Madry et~al.(2018)Madry, Makelov, Schmidt, Tsipras, and
  Vladu]{madry2018towards}
Aleksander Madry, Aleksandar Makelov, Ludwig Schmidt, Dimitris Tsipras, and
  Adrian Vladu.
\newblock Towards deep learning models resistant to adversarial attacks.
\newblock In \emph{International Conference on Learning Representations}, 2018.
\newblock URL \url{https://openreview.net/forum?id=rJzIBfZAb}.

\bibitem[Meinke and Hein(2020)]{Meinke2020Towards}
Alexander Meinke and Matthias Hein.
\newblock Towards neural networks that provably know when they don't know.
\newblock In \emph{International Conference on Learning Representations}, 2020.
\newblock URL \url{https://openreview.net/forum?id=ByxGkySKwH}.

\bibitem[Mirman et~al.(2018)Mirman, Gehr, and Vechev]{pmlr-v80-mirman18b}
Matthew Mirman, Timon Gehr, and Martin Vechev.
\newblock Differentiable abstract interpretation for provably robust neural
  networks.
\newblock In Jennifer Dy and Andreas Krause, editors, \emph{Proceedings of the
  35th International Conference on Machine Learning}, volume~80 of
  \emph{Proceedings of Machine Learning Research}, pages 3578--3586,
  Stockholmsmässan, Stockholm Sweden, 10--15 Jul 2018. PMLR.
\newblock URL \url{http://proceedings.mlr.press/v80/mirman18b.html}.

\bibitem[Mohapatra et~al.(2020)Mohapatra, Ko, Weng, Chen, Liu, and
  Daniel]{mohapatra2020higher}
Jeet Mohapatra, Ching-Yun Ko, Tsui-Wei Weng, Pin-Yu Chen, Sijia Liu, and Luca
  Daniel.
\newblock Higher-order certification for randomized smoothing.
\newblock In \emph{Advances in Neural Information Processing Systems}, 2020.

\bibitem[Moosavi-Dezfooli et~al.(2016)Moosavi-Dezfooli, Fawzi, and
  Frossard]{moosavi2016deepfool}
Seyed-Mohsen Moosavi-Dezfooli, Alhussein Fawzi, and Pascal Frossard.
\newblock Deep{F}ool: a simple and accurate method to fool deep neural
  networks.
\newblock In \emph{Proceedings of the IEEE Conference on Computer Vision and
  Pattern Recognition}, pages 2574--2582, 2016.

\bibitem[Pereyra et~al.(2017)Pereyra, Tucker, Chorowski, Kaiser, and
  Hinton]{pereyra2017regularizing}
Gabriel Pereyra, George Tucker, Jan Chorowski, {\L}ukasz Kaiser, and Geoffrey
  Hinton.
\newblock Regularizing neural networks by penalizing confident output
  distributions.
\newblock \emph{arXiv preprint arXiv:1701.06548}, 2017.

\bibitem[Qin et~al.(2019)Qin, Carlini, Cottrell, Goodfellow, and
  Raffel]{qin2019speech}
Yao Qin, Nicholas Carlini, Garrison Cottrell, Ian Goodfellow, and Colin Raffel.
\newblock Imperceptible, robust, and targeted adversarial examples for
  automatic speech recognition.
\newblock In \emph{Proceedings of the 36th International Conference on Machine
  Learning}, volume~97 of \emph{Proceedings of Machine Learning Research},
  pages 5231--5240, Long Beach, California, USA, 09--15 Jun 2019. PMLR.
\newblock URL \url{http://proceedings.mlr.press/v97/qin19a.html}.

\bibitem[Rice et~al.(2020)Rice, Wong, and Kolter]{pmlr-v119-rice20a}
Leslie Rice, Eric Wong, and Zico Kolter.
\newblock Overfitting in adversarially robust deep learning.
\newblock In Hal~Daumé III and Aarti Singh, editors, \emph{Proceedings of the
  37th International Conference on Machine Learning}, volume 119 of
  \emph{Proceedings of Machine Learning Research}, pages 8093--8104. PMLR,
  13--18 Jul 2020.
\newblock URL \url{http://proceedings.mlr.press/v119/rice20a.html}.

\bibitem[Russakovsky et~al.(2015)Russakovsky, Deng, Su, Krause, Satheesh, Ma,
  Huang, Karpathy, Khosla, Bernstein, Berg, and Fei-Fei]{dataset/ilsvrc}
Olga Russakovsky, Jia Deng, Hao Su, Jonathan Krause, Sanjeev Satheesh, Sean Ma,
  Zhiheng Huang, Andrej Karpathy, Aditya Khosla, Michael Bernstein,
  Alexander~C. Berg, and Li~Fei-Fei.
\newblock {ImageNet Large Scale Visual Recognition Challenge}.
\newblock \emph{International Journal of Computer Vision}, 115\penalty0
  (3):\penalty0 211--252, 2015.
\newblock \doi{10.1007/s11263-015-0816-y}.

\bibitem[Salman et~al.(2019)Salman, Li, Razenshteyn, Zhang, Zhang, Bubeck, and
  Yang]{nips_salman19}
Hadi Salman, Jerry Li, Ilya Razenshteyn, Pengchuan Zhang, Huan Zhang, Sebastien
  Bubeck, and Greg Yang.
\newblock Provably robust deep learning via adversarially trained smoothed
  classifiers.
\newblock In \emph{Advances in Neural Information Processing Systems 32}, pages
  11289--11300. Curran Associates, Inc., 2019.

\bibitem[Santurkar et~al.(2019)Santurkar, Ilyas, Tsipras, Engstrom, Tran, and
  Madry]{santurkar2019cvrobust}
Shibani Santurkar, Andrew Ilyas, Dimitris Tsipras, Logan Engstrom, Brandon
  Tran, and Aleksander Madry.
\newblock Image synthesis with a single (robust) classifier.
\newblock In \emph{Advances in Neural Information Processing Systems},
  volume~32, pages 1262--1273. Curran Associates, Inc., 2019.
\newblock URL
  \url{https://proceedings.neurips.cc/paper/2019/file/6f2268bd1d3d3ebaabb04d6b5d099425-Paper.pdf}.

\bibitem[Schmidt et~al.(2018)Schmidt, Santurkar, Tsipras, Talwar, and
  Madry]{schmidt2018adversarially}
Ludwig Schmidt, Shibani Santurkar, Dimitris Tsipras, Kunal Talwar, and
  Aleksander Madry.
\newblock Adversarially robust generalization requires more data.
\newblock In \emph{Advances in Neural Information Processing Systems}, 2018.

\bibitem[Stutz et~al.(2020)Stutz, Hein, and Schiele]{pmlr-v119-stutz20a}
David Stutz, Matthias Hein, and Bernt Schiele.
\newblock Confidence-calibrated adversarial training: Generalizing to unseen
  attacks.
\newblock In Hal~Daumé III and Aarti Singh, editors, \emph{Proceedings of the
  37th International Conference on Machine Learning}, volume 119 of
  \emph{Proceedings of Machine Learning Research}, pages 9155--9166. PMLR,
  13--18 Jul 2020.
\newblock URL \url{http://proceedings.mlr.press/v119/stutz20a.html}.

\bibitem[Szegedy et~al.(2014)Szegedy, Zaremba, Sutskever, Bruna, Erhan,
  Goodfellow, and Fergus]{szegedy2013intriguing}
Christian Szegedy, Wojciech Zaremba, Ilya Sutskever, Joan Bruna, Dumitru Erhan,
  Ian Goodfellow, and Rob Fergus.
\newblock Intriguing properties of neural networks.
\newblock In \emph{International Conference on Learning Representations}, 2014.

\bibitem[Tack et~al.(2020)Tack, Mo, Jeong, and Shin]{tack2020csi}
Jihoon Tack, Sangwoo Mo, Jongheon Jeong, and Jinwoo Shin.
\newblock {CSI}: Novelty detection via contrastive learning on distributionally
  shifted instances.
\newblock In H.~Larochelle, M.~Ranzato, R.~Hadsell, M.~F. Balcan, and H.~Lin,
  editors, \emph{Advances in Neural Information Processing Systems}, volume~33,
  pages 11839--11852. Curran Associates, Inc., 2020.
\newblock URL
  \url{https://proceedings.neurips.cc/paper/2020/file/8965f76632d7672e7d3cf29c87ecaa0c-Paper.pdf}.

\bibitem[Tramer et~al.(2020)Tramer, Carlini, Brendel, and
  Madry]{tramer2020adaptive}
Florian Tramer, Nicholas Carlini, Wieland Brendel, and Aleksander Madry.
\newblock On adaptive attacks to adversarial example defenses.
\newblock In \emph{Advances in Neural Information Processing Systems},
  volume~33, 2020.

\bibitem[Tsipras et~al.(2019)Tsipras, Santurkar, Engstrom, Turner, and
  Madry]{tsipras2018robustness}
Dimitris Tsipras, Shibani Santurkar, Logan Engstrom, Alexander Turner, and
  Aleksander Madry.
\newblock Robustness may be at odds with accuracy.
\newblock In \emph{International Conference on Learning Representations}, 2019.
\newblock URL \url{https://openreview.net/forum?id=SyxAb30cY7}.

\bibitem[Verma et~al.(2019)Verma, Lamb, Beckham, Najafi, Mitliagkas, Lopez-Paz,
  and Bengio]{pmlr-v97-verma19a}
Vikas Verma, Alex Lamb, Christopher Beckham, Amir Najafi, Ioannis Mitliagkas,
  David Lopez-Paz, and Yoshua Bengio.
\newblock Manifold mixup: Better representations by interpolating hidden
  states.
\newblock In Kamalika Chaudhuri and Ruslan Salakhutdinov, editors,
  \emph{Proceedings of the 36th International Conference on Machine Learning},
  volume~97 of \emph{Proceedings of Machine Learning Research}, pages
  6438--6447. PMLR, 09--15 Jun 2019.
\newblock URL \url{http://proceedings.mlr.press/v97/verma19a.html}.

\bibitem[Wang et~al.(2021)Wang, Zhai, He, Wang, and
  Jian]{wang2021pretraintofinetune}
Lei Wang, Runtian Zhai, Di~He, Liwei Wang, and Li~Jian.
\newblock Pretrain-to-finetune adversarial training via sample-wise randomized
  smoothing, 2021.
\newblock URL \url{https://openreview.net/forum?id=Te1aZ2myPIu}.

\bibitem[Wong and Kolter(2018)]{pmlr-v80-wong18a}
Eric Wong and Zico Kolter.
\newblock Provable defenses against adversarial examples via the convex outer
  adversarial polytope.
\newblock In Jennifer Dy and Andreas Krause, editors, \emph{Proceedings of the
  35th International Conference on Machine Learning}, volume~80 of
  \emph{Proceedings of Machine Learning Research}, pages 5286--5295,
  Stockholmsmässan, Stockholm Sweden, 10--15 Jul 2018. PMLR.
\newblock URL \url{http://proceedings.mlr.press/v80/wong18a.html}.

\bibitem[Xiao et~al.(2018)Xiao, Zhu, Li, He, Liu, and Song]{xiao2018spatially}
Chaowei Xiao, Jun-Yan Zhu, Bo~Li, Warren He, Mingyan Liu, and Dawn Song.
\newblock Spatially transformed adversarial examples.
\newblock In \emph{International Conference on Learning Representations}, 2018.
\newblock URL \url{https://openreview.net/forum?id=HyydRMZC-}.

\bibitem[Xiao et~al.(2019)Xiao, Tjeng, Shafiullah, and Madry]{xiao2018training}
Kai~Y. Xiao, Vincent Tjeng, Nur Muhammad~(Mahi) Shafiullah, and Aleksander
  Madry.
\newblock Training for faster adversarial robustness verification via inducing
  {ReLU} stability.
\newblock In \emph{International Conference on Learning Representations}, 2019.
\newblock URL \url{https://openreview.net/forum?id=BJfIVjAcKm}.

\bibitem[Yang et~al.(2020)Yang, Duan, Hu, Salman, Razenshteyn, and
  Li]{pmlr-v119-yang20c}
Greg Yang, Tony Duan, J.~Edward Hu, Hadi Salman, Ilya Razenshteyn, and Jerry
  Li.
\newblock Randomized smoothing of all shapes and sizes.
\newblock In Hal~Daumé III and Aarti Singh, editors, \emph{Proceedings of the
  37th International Conference on Machine Learning}, volume 119 of
  \emph{Proceedings of Machine Learning Research}, pages 10693--10705. PMLR,
  13--18 Jul 2020.
\newblock URL \url{http://proceedings.mlr.press/v119/yang20c.html}.

\bibitem[Yun et~al.(2019)Yun, Han, Oh, Chun, Choe, and Yoo]{yun2019cutmix}
Sangdoo Yun, Dongyoon Han, Seong~Joon Oh, Sanghyuk Chun, Junsuk Choe, and
  Youngjoon Yoo.
\newblock {CutMix}: Regularization strategy to train strong classifiers with
  localizable features.
\newblock In \emph{Proceedings of the IEEE/CVF International Conference on
  Computer Vision}, pages 6023--6032, 2019.

\bibitem[Yurtsever et~al.(2020)Yurtsever, Lambert, Carballo, and
  Takeda]{yurtsever2020survey}
Ekim Yurtsever, Jacob Lambert, Alexander Carballo, and Kazuya Takeda.
\newblock A survey of autonomous driving: Common practices and emerging
  technologies.
\newblock \emph{IEEE Access}, 8:\penalty0 58443--58469, 2020.

\bibitem[Zhai et~al.(2020)Zhai, Dan, He, Zhang, Gong, Ravikumar, Hsieh, and
  Wang]{Zhai2020MACER}
Runtian Zhai, Chen Dan, Di~He, Huan Zhang, Boqing Gong, Pradeep Ravikumar,
  Cho-Jui Hsieh, and Liwei Wang.
\newblock {MACER}: Attack-free and scalable robust training via maximizing
  certified radius.
\newblock In \emph{International Conference on Learning Representations}, 2020.
\newblock URL \url{https://openreview.net/forum?id=rJx1Na4Fwr}.

\bibitem[Zhang et~al.(2019)Zhang, Yu, Jiao, Xing, Ghaoui, and
  Jordan]{pmlr-v97-zhang19p}
Hongyang Zhang, Yaodong Yu, Jiantao Jiao, Eric Xing, Laurent~El Ghaoui, and
  Michael Jordan.
\newblock Theoretically principled trade-off between robustness and accuracy.
\newblock In \emph{Proceedings of the 36th International Conference on Machine
  Learning}, volume~97 of \emph{Proceedings of Machine Learning Research},
  pages 7472--7482, Long Beach, California, USA, 09--15 Jun 2019. PMLR.
\newblock URL \url{http://proceedings.mlr.press/v97/zhang19p.html}.

\bibitem[Zhang et~al.(2018)Zhang, Cisse, Dauphin, and
  Lopez-Paz]{zhang2018mixup}
Hongyi Zhang, Moustapha Cisse, Yann~N. Dauphin, and David Lopez-Paz.
\newblock mixup: Beyond empirical risk minimization.
\newblock In \emph{International Conference on Learning Representations}, 2018.
\newblock URL \url{https://openreview.net/forum?id=r1Ddp1-Rb}.

\bibitem[Zhang et~al.(2020{\natexlab{a}})Zhang, Chen, Xiao, Gowal, Stanforth,
  Li, Boning, and Hsieh]{zhang2020towards}
Huan Zhang, Hongge Chen, Chaowei Xiao, Sven Gowal, Robert Stanforth, Bo~Li,
  Duane Boning, and Cho-Jui Hsieh.
\newblock Towards stable and efficient training of verifiably robust neural
  networks.
\newblock In \emph{International Conference on Learning Representations},
  2020{\natexlab{a}}.
\newblock URL \url{https://openreview.net/forum?id=Skxuk1rFwB}.

\bibitem[Zhang et~al.(2020{\natexlab{b}})Zhang, Xu, Han, Niu, Cui, Sugiyama,
  and Kankanhalli]{pmlr-v119-zhang20z}
Jingfeng Zhang, Xilie Xu, Bo~Han, Gang Niu, Lizhen Cui, Masashi Sugiyama, and
  Mohan Kankanhalli.
\newblock Attacks which do not kill training make adversarial learning
  stronger.
\newblock In \emph{Proceedings of the 37th International Conference on Machine
  Learning}, volume 119 of \emph{Proceedings of Machine Learning Research},
  pages 11278--11287. PMLR, 13--18 Jul 2020{\natexlab{b}}.
\newblock URL \url{http://proceedings.mlr.press/v119/zhang20z.html}.

\bibitem[Zhang et~al.(2021)Zhang, Deng, Kawaguchi, Ghorbani, and
  Zou]{zhang2021how}
Linjun Zhang, Zhun Deng, Kenji Kawaguchi, Amirata Ghorbani, and James Zou.
\newblock How does mixup help with robustness and generalization?
\newblock In \emph{International Conference on Learning Representations}, 2021.
\newblock URL \url{https://openreview.net/forum?id=8yKEo06dKNo}.

\end{thebibliography}
\clearpage
\appendix

\FloatBarrier
\section{Training procedure of SmoothMix}
\label{ap:alg}

\begin{algorithm}[ht]
\caption{SmoothMix training}\label{alg:training}
\begin{algorithmic}[1]
\REQUIRE Sample $(x, y)\sim P$. smoothing factor $\sigma$. number of noise samples $m$. number of steps $T$. step size $\alpha$. regularization strength $\eta > 0$. 
\vspace{0.05in}
\hrule
\vspace{0.05in}
\STATE Sample $\delta_1, \cdots, \delta_m \sim \mathcal{N}(0, \sigma^2 I)$, and $\lambda \sim \mathcal{U}(\left[0, \tfrac{1}{2}\right])$
\STATE {{\textsc{// Find an adversarial example}}}
\STATE $\tilde{x}^{(0)}, \hat{F}(x^{(0)}) \leftarrow x, \frac{1}{m}\sum_{i=1}^m F(x+\delta_i)$
\FOR{$t=0$ {\bfseries to} $T - 1$}
    \STATE $J(\tilde{x}^{(t)}) \leftarrow -\log \hat{F}_y(\tilde{x}^{(t)})$
    \STATE $\tilde{x}^{(t+1)} \leftarrow \tilde{x}^{(t)} + \alpha \cdot \frac{\nabla_x J(\tilde{x}^{(t)})}{\|\nabla_x J(\tilde{x}^{(t)})\|_2}$
    \STATE $\hat{F}(\tilde{x}^{(t+1)}) \leftarrow \frac{1}{m}\sum_{i=1}^m F(\tilde{x}^{(t+1)}+\delta_i)$
\ENDFOR
\SHORTIF{$~\mathtt{use\_single\_step}~$}{$~x \leftarrow \tilde{x}^{(1)}~$} 
\STATE {{\textsc{// Compute the SmoothMix loss}}}
\STATE $x^{\tt mix}, y^{\tt mix} \leftarrow ((1-\lambda) \cdot x + \lambda \cdot \tilde{x}^{(T)}), ((1 - \lambda) \cdot \hat{F}(x) + \lambda \cdot \tfrac{\mathbbm{1}}{C})$
\FOR{$i=1$ {\bfseries to} $m$}
    \STATE $L^{\tt nat}_{i}, L^{\tt mix}_{i} \leftarrow \mathcal{L}(F(x+\delta_i), y), \mathcal{L}(F(x^{\tt mix}+\delta_i), y^{\tt mix})$
\ENDFOR
\STATE $L \leftarrow \frac{1}{m}\sum_i (L^{\tt nat}_{i} + \eta \cdot L^{\tt mix}_{i})$
\end{algorithmic}
\end{algorithm}

\FloatBarrier
\section{Discussion on input-dependent designs of noise scales}
\label{ap:adaptive_sigma}

\jh{In this paper, we aim to develop a new training method to exhibit a better trade-off between accuracy and (certified) robustness of smoothed classifiers. Meanwhile, there has been recently another proposal to improve the robustness of smoothed classifiers without a new training scheme, namely by certifying a given smoothed classifier with \emph{input-dependent} $\sigma$ \cite{alfarra2020data, wang2021pretraintofinetune, chen2021instars}.}
In this section, however, we show that if one allows \emph{different} noise scales $\sigma$ for each input in attempt to generalize the current framework of randomized smoothing \cite{pmlr-v97-cohen19c}, then the actual robustness guarantee would rapidly decrease as the input dimension grows. In particular, we consider the following classifier $\tilde f$ generalizing \eqref{eq:smoothing} with some non-negative function $g:\mathbb R^d\rightarrow\mathbb R_{\ge0}$, defined as follows:
\begin{align*}
    \tilde f(x):=\argmax_{c\in\mathcal Y}\mathbb P_{\delta\sim\mathcal N(0,g(x)I)}(f(x+\delta)=c),
\end{align*}
In other words, we assume that the scaling parameter of the smoothing noise can now be a function of $x$. As in the main text, we are interested in the certified radius $\underline R(\tilde f;x,y)$ of $\tilde f$.

One may expect that $\underline R(\tilde f;x,y)$ can be significantly larger than $\underline R(\hat f;x,y)$ since $\hat f$ is a special case of $\tilde f$, \ie constant $g(x)$. However, we show that it may not be true for high-dimensional inputs: even a small deviation of $g(x)$ can incur very poor certified robustness.
Formally, we prove the following theorem. 
\begin{theorem}\label{thm:dimdep}
Let $r_i,i\in\mathbb N$ be any i.i.d.\ random variables of zero mean, unit variance, and $\mathbb E[r_i^4]<\infty$. Let $\mathcal F_d$ be a collection of all measurable functions from $\mathbb R^d$ to $\{0,1\}$.
Let $p\in(0.5,1)$, $\sigma,\tau>0$, and $\varepsilon\in(0,1/2]$ be constants such that $\sigma\ne\tau$.
Then, for $\delta:=(r_1,\dots,r_d)$, for any $c\in\{0,1\}$, and for any $d\in\mathbb N$, the following statements hold: 
\begin{align*}
\sup_{x,x^\prime\in\mathbb R^d:\|x-x^\prime\|_2\le \varepsilon}\inf_{f\in\mathcal F_d:\mathbb P(f(x+\sigma\delta)=c)=p}\mathbb P(f(x^\prime+\tau\delta)=c)\le C/d.
\end{align*}
for some constant $C>0$ which is a function of other constants $p,\sigma,\tau,\varepsilon,\mathbb E[r_i^4]$.
\end{theorem}
Theorem \ref{thm:dimdep} indicates the curse of dimensionality for the worst classifier under general noises of a finite kurtosis. In particular, it states that there exists an upper bound on $\mathbb P(f(x+\tau\delta)=c)$ inversely proportional to the input dimension $d$ even though two inputs $x,x^\prime$ are extremely close. Hence, if we utilize different noise scales (\ie $\sigma$) for each input, the resulting lower bound on the certified radius relying on the worst-case bound as in   \citep{pmlr-v97-cohen19c,lecuyer2019certified,nips_salman19} will be small for high-dimensional inputs. Namely, choosing (almost) constant noise scale for the inputs in the target certification region is necessary.

\subsection{Proof of Theorem \ref{thm:dimdep}}\label{ss:pfthm:dimdep}
We first define $z=(z_1,\dots,z_d):=x^\prime-x$, \ie $\|z\|_2\le\varepsilon$. Then,
the following inequality trivially holds.
\begin{align}
    \inf_{f\in\mathcal F_d:\mathbb E[f(x+\sigma\delta)]=p}\mathbb E[f(x^\prime+\tau\delta)]&\le\inf_{\mathcal U\subset\mathbb R^d:\mathbb P(\sigma\delta\in\mathcal U)=p}\mathbb P\left(\tau \delta+z\in\mathcal U\right)\notag\\
    &\le\mathbb P\left(\frac{\|\tau\delta+z\|_2^2}d\in[\sigma^2-k,\sigma^2+k]\right)\label{eq:pfthmdimdep}
\end{align}
where $k$ is a non-negative number satisfying 
\begin{align*}
    \mathbb P\left(\frac{\|\sigma\delta\|_2^2}d\in[\sigma^2-k,\sigma^2+k]\right)=p.
\end{align*}
The following lemma asserts that the RHS of \eqref{eq:pfthmdimdep} is bounded by $C/d$ where $C$ is some constant which is only a function of $\mathbb E[r_i^4],\sigma,\tau,\varepsilon,p$. This completes the proof of Theorem \ref{thm:dimdep}.
\begin{lemma}\label{lem:dimdependence3}
There exists $C$ which is a function of $\mathbb E[r_i^4],\sigma,\tau,\varepsilon,p$ such that the following statements hold: for any $d\in\mathbb N$ and for any $z\in\mathbb R^d$ satisfying $\|z\|_2\le\varepsilon$,
$$\mathbb P\left(\frac{\|\tau\delta+z\|_2^2}d\in[\sigma^2-k,\sigma^2+k]\right)\le \frac{C}d.$$
\end{lemma}

\subsection{Proof of Lemma \ref{lem:dimdependence3}}
Lemma \ref{lem:dimdependence3} is a direct consequence of the law of large numbers applied to the i.i.d.\ random variables $r_i^2$. 
First, we compute the variance of $\frac{\|\sigma\delta\|_2^2}d$ using the following equality: for $\eta:=\sqrt{\mathbb E[r_i^4]-1}$,
\begin{align*}
    \text{Var}\left(\frac{\|\sigma\delta\|_2^2}d\right)&=\mathbb E\left[\left(\frac{\|\sigma\delta\|_2^2}d-\sigma^2\right)^2\right]=\mathbb E\left[\left(\frac{\sigma^2}{d}\sum_{i=1}^d(r_i^2-1)\right)^2\right]
    \\
    &=\frac{\sigma^4}{d^2}\sum_{i=1}^d\mathbb E\left[(r_i^2-1)^2\right]=\frac{\sigma^4}{d^2}\sum_{i=1}^d\mathbb E[r_i^4]-1\\
    &=\frac{\sigma^4(\mathbb E[r_i^4]-1)}d=\frac{\sigma^4\eta^2}{d}
\end{align*}
where the third equality follows from the independence of $r_i$s and the fourth inequality follows from $\mathbb E[r_i^2]=1$.
Hence, from the Chebyshev's inequality, we have
\begin{align}
    \mathbb P\left(\left|\frac{\|\sigma\delta\|_2^2}{d}-\sigma^2\right|<\frac{\sigma^2\eta}{\sqrt{d(1-p)}}\right)\ge1-(\sqrt{1-p})^2=p,\label{eq:dimdeplem1}
\end{align}
\ie $k\le\frac{\sigma^2\eta}{\sqrt{d(1-p)}}$.

Now, we derive a similar concentration inequality for $\frac{\|\tau\delta+z\|_2^2}d$. 
To this end, we bound its deviation from $\tau^2+\frac{\|z\|_2^2}{d}$ as follows:
\begin{align}
    &\mathbb P\left(\left|\frac{\|\tau\delta+z\|_2^2}{d}-\left(\tau^2+\frac{\|z\|_2^2}d\right)\right|\ge\frac{\sigma^2+\tau^2}3\right)\notag\\
    &=1-\mathbb P\left(\left|\frac{\|\tau\delta+z\|_2^2}{d}-\left(\tau^2+\frac{\|z\|_2^2}d\right)\right|<\frac{\sigma^2+\tau^2}3\right)\notag\\
    &\le1-\mathbb P\left(\left|\frac{\|\tau\delta\|_2^2}{d}-\tau^2\right|<\frac{\sigma^2+\tau^2}6\quad\text{and}\quad\left|\frac{2\tau\sum_{i=1}^dr_iz_i}{d}\right|<\frac{\sigma^2+\tau^2}6\right)\notag\\
    &=\mathbb P\left(\left|\frac{\|\tau\delta\|_2^2}{d}-\tau^2\right|\ge\frac{\sigma^2+\tau^2}6\quad\text{or}\quad\left|\frac{2\tau\sum_{i=1}^dr_iz_i}{d}\right|\ge\frac{\sigma^2+\tau^2}6\right)\notag\\
    &\le\mathbb P\left(\left|\frac{\|\tau\delta\|_2^2}{d}-\tau^2\right|\ge\frac{\sigma^2+\tau^2}6\right)+\mathbb P\left(\left|\frac{2\tau\sum_{i=1}^dr_iz_i}{d}\right|\ge\frac{\sigma^2+\tau^2}6\right)\notag\\
    &\le\frac{36\tau^4\eta^2+144\tau^2\varepsilon^2}{(\sigma^2+\tau^2)^2d}\label{eq:dimdeplem2}
\end{align}
where the last inequality is from the variance bounds
\begin{align*}
&\text{Var}\left(\frac{\|\tau\delta\|_2^2}{d}-\tau^2\right)=\frac{\tau^4\eta^2}{d}\\
&\text{Var}\left(\frac{2\tau\sum_{i=1}^dr_iz_i}{d}\right)=\frac{4\tau^2\|z\|_2^2}{d^2}\le\frac{4\tau^2\varepsilon^2}{d^2}
\end{align*}
and the Chebyshev's inequality
\begin{align*}
&\mathbb P\left(\left|\frac{\|\tau\delta\|_2^2}{d}-\tau^2\right|\ge \frac{\sigma^2+\tau^2}{6}\right)\le\frac{36\tau^4\eta^2}{(\sigma^2+\tau^2)^2d}\\
&\mathbb P\left(\left|\frac{2\tau\sum_{i=1}^dr_iz_i}{d}\right|\ge\frac{\sigma^2+\tau^2}6\right)\le\frac{144\tau^2\varepsilon^2}{(\sigma^2+\tau^2)^2d^2}.
\end{align*}
Then, for all $d\ge\max\left\{\frac{4\sigma^4\eta^2}{(\tau^2-\sigma^2)^2(1-p)},\frac{6\varepsilon^2}{\sigma^2+\tau^2}\right\}$, \ie $\frac{\sigma^2\eta}{\sqrt{d(1-p)}}\le\frac{|\tau^2-\sigma^2|}{2}$ and $\frac{\varepsilon^2}d\le\frac{\sigma^2+\tau^2}6$, it holds that
\begin{align*}
    &\mathbb P\left(\frac{\|\tau\delta+z\|_2^2}d\in[\sigma^2-k,\sigma^2+k]\right)\\
    &\le\mathbb P\left(\frac{\|\tau\delta+z\|_2^2}d\in\left[\sigma^2-\frac{|\tau^2-\sigma^2|}{2},\sigma^2+\frac{|\tau^2-\sigma^2|}{2}\right]\right)\\
    &\le\mathbb P\left(\left|\frac{\|\tau\delta+z\|_2^2}{d}-\left(\tau^2+\frac{\|z\|_2^2}d\right)\right|\ge\frac{\sigma^2+\tau^2}3\right)\\
    &\le\frac{36\tau^4\eta^2+144\tau^2\varepsilon^2}{(\sigma^2+\tau^2)^2d}
\end{align*}
by using \eqref{eq:dimdeplem2}. Hence, choosing
\begin{align*}
C:=\max\left\{\frac{4\sigma^4\eta^2}{(\tau^2-\sigma^2)^2(1-p)},\frac{6\varepsilon^2}{\sigma^2+\tau^2},\frac{36\tau^4\eta^2+144\tau^2\varepsilon^2}{(\sigma^2+\tau^2)^2}\right\}
\end{align*}
completes the proof of Lemma \ref{lem:dimdependence3}.

\FloatBarrier
\clearpage
\section{Experimental details}
\label{ap:details}

Throughout our experiments, we follow the same training details of prior works \cite{pmlr-v97-cohen19c, nips_salman19, Zhai2020MACER, jeong2020consistency} for a fair comparison: more specifically, we use LeNet \cite{dataset/mnist} for MNIST, ResNet-110 \cite{he2016deep} for CIFAR-10, and ResNet-50 \cite{he2016deep} for ImageNet.
We train every model via stochastic gradient descent using Nesterov momentum of weight 0.9 without dampening. We set a weight decay of $10^{-4}$ for all the models. 
We consider three different noise levels $\sigma\in\{0.25, 0.5, 1.0\}$ for smoothing classifiers for MNIST and CIFAR-10 models, and $\sigma\in\{0.5, 1.0\}$ in the case of ImageNet. 
We used up to 4 NVIDIA TITAN Xp GPUs to run each configurations considered in our experiments, both for training and certification: more specifically, we used a single GPU to run every experimenet on MNIST and CIFAR-10, and four GPUs to run ImageNet models. 

\subsection{Datasets}

\textbf{MNIST} dataset \citep{dataset/mnist} consists 70,000 gray-scale hand-written digit images of size 28$\times$28, 60,000 for training and 10,000 for testing. Each of the images is labeled from 0 to 9, i.e., there are 10 classes. 
We do not perform any pre-processing except for normalizing the range of each pixel from 0-255 to 0-1. 
When MNIST is used for training, we use LeNet \cite{dataset/mnist} for 90 epochs and use the initial learning rate of 0.01. The learning rate is decayed by 0.1 at 30-th and 60-th epoch.

\textbf{CIFAR-10} dataset \citep{dataset/cifar} consist of 60,000 RGB images of size 32$\times$32 pixels, 50,000 for training and 10,000 for testing. Each of the images is labeled to one of 10 classes, 
and the number of data per class is set evenly, i.e., 6,000 images per each class. 
We use the standard data-augmentation scheme of random horizontal flip and random translation up to 4 pixels, as also used by other baselines \cite{pmlr-v97-cohen19c, nips_salman19, Zhai2020MACER, jeong2020consistency}. 
We also normalize the images in pixel-wise by the mean and the standard deviation calculated from the training set. 
When CIFAR-10 is used for training, we train ResNet-110 \cite{he2016deep} models for 150 epochs with initial learning rate of 0.1. The learning rate if decated by 0.1 at 50-th and 100-th epoch. 

\textbf{ImageNet} classification dataset \cite{dataset/ilsvrc} consists of 1.2 million training images and 50,000 validation images, which are labeled by one of 1,000 classes. 
For data-augmentation, we perform 224$\times$224 random cropping with random resizing and horizontal flipping to the training images. At test time, on the other hand, 224$\times$224 center cropping is performed after re-scaling the images into 256$\times$256. 
When ImageNet is used for training, we train ResNet-50 \cite{he2016deep} models for 90 epochs with initial learning rate of 0.1. The learning rate if decated by 0.1 at 30-th and 60-th epoch.

\subsection{Detailed hyperparameters for baselines}
\label{ap:hyper}

\textbf{Stability training \cite{li2019stab}}  uses a single hyperparameter $\lambda > 0$ to control the relative strength of the stability regularization compared to the standard cross-entropy loss. In our experiments, we use $\lambda=2$ by default for this method, but except for the ``$\sigma=1.0$'' model on CIFAR-10: in this case, we had to reduce it to $\lambda=1$ for a stable training.

\textbf{SmoothAdv \cite{nips_salman19}} mainly controls three hyperparameters those are for performing projected gradient descent (PGD) to find adversarial examples in the training: namely, it uses $m$: the number of noise samples, $T$: the number of PGD steps, and $\varepsilon$: an $\ell_2$-norm restriction on adversarial perturbations. For SmoothAdv models, we fix $T=10$ and $\varepsilon=1.0$ throughout the experiments. In case of $m$, and use $m=4$ for MNIST models, and $m=8$ for CIFAR-10. Following \citet{nips_salman19}, we also adopt the \emph{warm-up} strategy on $\varepsilon$, \ie it is initially set to zero, and gradually increased for the first 10 epochs up to the original value of $\varepsilon$.

\textbf{MACER \cite{Zhai2020MACER}} adds four hyperparameters to the training: namely, it uses $m$: the number of noise samples, $\lambda$: the relative strength of regularization, $\beta$: a temperature scaling factor, and $\gamma$: a margin gap. We follow the configurations reported by \citet{Zhai2020MACER} to reproduce the MNIST results: namely, we use $m=16$, $\beta=16.0$, $\gamma=8.0$ and $\lambda=16.0$. 
We use $\lambda=6.0$ in case of $\sigma=1.0$ on MNIST, however, for a better training stability. We use the pre-trained models released by the authors for evaluations on CIFAR-10, which can be downloaded at \url{https://github.com/RuntianZ/macer}. These CIFAR-10 models are reported to be trained with $m=16$, $\beta=16.0$, $\gamma=8.0$, and $\lambda=12.0$ and $4.0$ for $\sigma=0.25$ and $0.5$, respectively. For $\sigma=1.0$, $\lambda$ is initially set to 0, and changed to $\lambda=12.0$ after the first learning rate decay.

\textbf{Consistency \cite{jeong2020consistency}} controls two hyperparameters, namely $\lambda$ and $\eta$, each for the relative strength of the consistency term and the entropy term, respectively. We obtain results from the best hyperparameters those reported by \citet{jeong2020consistency} when the consistency regularization is applied to the Gaussian training baseline, both in MNIST and CIFAR-10 datasets. More concretely, we fix $\eta=0.5$ for every model, and use $\lambda=5$ for MNIST and $\lambda=10$ for CIFAR-10 models by default. In case of $\sigma=0.25$, $\lambda$ is doubled in both datasets, \ie $\lambda=10$ and $\lambda=20$ for MNIST and CIFAR-10, respectively, as it is shown to achieve better ACRs.

\section{Related work}

\textbf{Certified adversarial robustness. }
We focus on improving adversarial robustness of \emph{randomized smoothing} \cite{pmlr-v97-cohen19c} based classifiers, which is currently one of prominent ways to obtain a classifier with a robustness certification. In general, there have been many attempts other than randomized smoothing to provide a robustness certification of deep neural networks \cite{gehr2018ai2, pmlr-v80-wong18a, pmlr-v80-mirman18b, xiao2018training, gowal2019scalable, zhang2020towards}, and correspondingly with attempts to further improve the robustness with respect to those certification protocols \cite{pmlr-v89-croce19a, Croce2020Provable, Balunovic2020Adversarial}. Nevertheless, randomized smoothing has attracted particular attention as the first approach that could successfully scaled up to the ImageNet dataset \cite{dataset/ilsvrc}. 
A more complete taxonomy on the literature can be found in \citet{li2020sokcertified}.

\textbf{Confidence-calibrated training. }
\emph{Overconfident predictions} of deep neural networks \cite{pereyra2017regularizing} have been considered as problematic in many scenarios, \eg uncertainty estimation of in-distribution samples \cite{pmlr-v70-guo17a, NEURIPS2018_7180cffd, NEURIPS2019_f8c0c968}, those of out-of-distribution samples \cite{hendrycks2016baseline, lee2018training, Meinke2020Towards}, and ensemble learning \cite{pmlr-v70-lee17b}, just to name a few. In the context of adversarial training, \citet{pmlr-v119-stutz20a} have shown that regularizing confidence on adversarial examples to be uniform can improve detection of adversarial examples from unseen threat models. 
In this paper, we address the overconfidence at adversarial examples particularly focusing on \emph{smoothed classifiers}, observing that a simple approach of directly fixing the problem could significantly improve the certified robustness. 

\textbf{Mixup-based training. }
Originally, \emph{mixup} \cite{zhang2018mixup} has proposed as a simple yet effective data augmentation scheme to improve generalization and robustness (against small adversarial attacks) of deep neural networks, and there have been significant follow-up works to further improve this form
\cite{pmlr-v97-verma19a, yun2019cutmix, pmlr-v119-kim20b, kim2021comixup}. 
Recently, \citet{zhang2021how} have also explored on theoretical justifications behind how could such an augmentation improves generalization and robustness. 
Although our method uses a similar linear interpolation scheme of mixup, 
there is still an essential difference between ours and this line of works: namely, we do not rely on the prior of interpolating two (or more) \emph{independent} samples, but rather aims to directly calibrate predictions between a clean and its (unrestricted) adversarial example, \ie we consider a new form of \emph{self-mixup} training.

There have been also attempts to employ mixup particularly for improving adversarial robustness: \citet{lamb2019interpolated} have shown that an additional mixup loss between adversarial examples upon the standard mixup training achieves a comparable robustness to adversarial training (AT) \cite{madry2018towards}, while not compromising the clean accuracy as much as AT; 
\citet{lee2020adversarial} have proposed \emph{Adversarial Vertex Mixup} to improve AT, by extrapolating predictions along the direction of adversarial perturbation up to few times of its norm via mixup training. 
Our proposed method can be differentiated to these approaches, in a sense that we employ mixup not to directly improve the robustness of a given neural network, but of its smoothed counterpart. It is also our unique perspective that we consider \emph{unrestricted} adversarial examples to be interpolated.

\FloatBarrier
\section{Variance of results over multiple runs}
\label{ap:multiple_run}
    In our experiments, we report single-run results for ACR and certified robust accuracy as also done by \cite{pmlr-v97-cohen19c, nips_salman19, li2019stab, Zhai2020MACER, jeong2020consistency}, considering that ACR is fairly a robust metric to network initialization: \eg
    in Table~\ref{tab:mnist_var}, we report the mean and standard deviation of ACRs across 5 seeds for the MNIST results reported in Table~\ref{tab:mnist}. Overall, we confirm that ACR generally shows low variance over multiple runs across a wide range of training methods, including ours.
    
    \begin{table*}[ht]
    \centering
    \caption{Comparison of certified test accuracy for various training methods on MNIST. The reported values are the mean and standard deviation across 5 seeds. We set our result bold-faced whenever the value improves the Gaussian baseline, and the underlined are best-performing model per $\sigma$.}
    \label{tab:mnist_errorbar}
        \vspace{0.03in}
        \small
        \begin{adjustbox}{width=1.0\linewidth}
        \begin{tabular}{cl|cccccccccccc}
    \toprule
    $\sigma$ &  Models (MNIST) & 0.00 & 0.50 & 1.00 & 1.50 & 2.00 & 2.50 \\ 
    \midrule
    \multirow{9.5}{*}{0.25}  & Gaussian \cite{pmlr-v97-cohen19c} & 99.25 $\pm$ 0.04 & 96.75 $\pm$ 0.11 & 0.00 $\pm$ 0.00 & 0.00 $\pm$ 0.00 & 0.00 $\pm$ 0.00 & 0.00 $\pm$ 0.00 \\ 
     & Stability training \cite{li2019stab} & 99.34 $\pm$ 0.04 & 97.12 $\pm$ 0.12 & 0.00 $\pm$ 0.00 & 0.00 $\pm$ 0.00 & 0.00 $\pm$ 0.00 & 0.00 $\pm$ 0.00 \\ 
     & SmoothAdv \cite{nips_salman19} & 99.39 $\pm$ 0.01 & 98.17 $\pm$ 0.06 & 0.00 $\pm$ 0.00 & 0.00 $\pm$ 0.00 & 0.00 $\pm$ 0.00 & 0.00 $\pm$ 0.00 \\ 
     & MACER \cite{Zhai2020MACER} & 99.33 $\pm$ 0.03 & 97.35 $\pm$ 0.08 & 0.00 $\pm$ 0.00 & 0.00 $\pm$ 0.00 & 0.00 $\pm$ 0.00 & 0.00 $\pm$ 0.00 \\ 
     & Consistency \cite{jeong2020consistency} & 99.43 $\pm$ 0.03 & 97.92 $\pm$ 0.09 & 0.00 $\pm$ 0.00 & 0.00 $\pm$ 0.00 & 0.00 $\pm$ 0.00 & 0.00 $\pm$ 0.00 \\ 
    \cmidrule(l){2-2} \cmidrule(l){3-8}
    & \textbf{SmoothMix} ($\eta$ = 1.0) & \textbf{99.43 $\pm$ 0.03} & \textbf{98.10 $\pm$ 0.06} & 0.00 $\pm$ 0.00 & 0.00 $\pm$ 0.00 & 0.00 $\pm$ 0.00 & 0.00 $\pm$ 0.00 \\ 
    & \textbf{+ One-Step adversary} & \textbf{99.39 $\pm$ 0.02} & \textbf{98.17 $\pm$ 0.06} & 0.00 $\pm$ 0.00 & 0.00 $\pm$ 0.00 & 0.00 $\pm$ 0.00 & 0.00 $\pm$ 0.00 \\ 
    & \textbf{SmoothMix} ($\eta$ = 5.0) & \textbf{\underline{99.45} $\pm$ 0.03} & \textbf{98.17 $\pm$ 0.07} & 0.00 $\pm$ 0.00 & 0.00 $\pm$ 0.00 & 0.00 $\pm$ 0.00 & 0.00 $\pm$ 0.00 \\ 
    & \textbf{+ One-Step adversary} & \textbf{99.37 $\pm$ 0.02} & \textbf{\underline{98.20} $\pm$ 0.03} & 0.00 $\pm$ 0.00 & 0.00 $\pm$ 0.00 & 0.00 $\pm$ 0.00 & 0.00 $\pm$ 0.00 \\ 
    \midrule
    \multirow{9.5}{*}{0.50}  & Gaussian \cite{pmlr-v97-cohen19c} & 99.15 $\pm$ 0.03 & 96.90 $\pm$ 0.06 & 89.83 $\pm$ 0.06 & 67.80 $\pm$ 0.16 & 0.00 $\pm$ 0.00 & 0.00 $\pm$ 0.00 \\ 
     & Stability training \cite{li2019stab} & 99.26 $\pm$ 0.02 & 97.27 $\pm$ 0.09 & 90.75 $\pm$ 0.11 & 69.15 $\pm$ 0.38 & 0.00 $\pm$ 0.00 & 0.00 $\pm$ 0.00 \\ 
     & SmoothAdv \cite{nips_salman19} & 99.03 $\pm$ 0.03 & 97.36 $\pm$ 0.06 & 92.94 $\pm$ 0.08 & 81.06 $\pm$ 0.12 & 0.00 $\pm$ 0.00 & 0.00 $\pm$ 0.00 \\ 
     & MACER \cite{Zhai2020MACER} & 98.69 $\pm$ 0.09 & 96.28 $\pm$ 0.17 & 90.14 $\pm$ 0.20 & 72.12 $\pm$ 0.75 & 0.00 $\pm$ 0.00 & 0.00 $\pm$ 0.00 \\ 
     & Consistency \cite{jeong2020consistency} & 99.15 $\pm$ 0.03 & 97.51 $\pm$ 0.07 & 92.89 $\pm$ 0.10 & 78.26 $\pm$ 0.23 & 0.00 $\pm$ 0.00 & 0.00 $\pm$ 0.00 \\
    \cmidrule(l){2-2} \cmidrule(l){3-8}
    & \textbf{SmoothMix} ($\eta$ = 1.0) & 99.10 $\pm$ 0.02 & \textbf{\underline{97.51} $\pm$ 0.07} & \textbf{92.91 $\pm$ 0.08} & \textbf{80.15 $\pm$ 0.05} & 0.00 $\pm$ 0.00 & 0.00 $\pm$ 0.00 \\ 
    & \textbf{+ One-Step adversary} & 98.74 $\pm$ 0.04 & \textbf{97.09 $\pm$ 0.06} & \textbf{92.67 $\pm$ 0.05} & \textbf{\underline{81.70} $\pm$ 0.05} & 0.00 $\pm$ 0.00 & 0.00 $\pm$ 0.00 \\ 
    & \textbf{SmoothMix} ($\eta$ = 5.0) & 98.64 $\pm$ 0.04 & \textbf{96.98 $\pm$ 0.02} & \textbf{92.63 $\pm$ 0.07} & \textbf{\underline{81.85} $\pm$ 0.10} & 0.00 $\pm$ 0.00 & 0.00 $\pm$ 0.00 \\ 
    & \textbf{+ One-Step adversary} & 98.21 $\pm$ 0.02 & 96.34 $\pm$ 0.04 & \textbf{91.46 $\pm$ 0.03} & \textbf{\underline{81.20} $\pm$ 0.15} & 0.00 $\pm$ 0.00 & 0.00 $\pm$ 0.00 \\
    \midrule
    \multirow{9.5}{*}{1.00}  & Gaussian \cite{pmlr-v97-cohen19c} & 96.34 $\pm$ 0.03 & 91.39 $\pm$ 0.05 & 79.86 $\pm$ 0.08 & 59.49 $\pm$ 0.10 & 32.46 $\pm$ 0.20 & 10.93 $\pm$ 0.12 \\ 
     & Stability training \cite{li2019stab} & 96.43 $\pm$ 0.05 & 91.63 $\pm$ 0.05 & 80.45 $\pm$ 0.16 & 60.53 $\pm$ 0.07 & 33.35 $\pm$ 0.13 & 11.05 $\pm$ 0.13 \\ 
     & SmoothAdv \cite{nips_salman19} & 95.76 $\pm$ 0.03 & 90.72 $\pm$ 0.07 & 80.81 $\pm$ 0.14 & 64.44 $\pm$ 0.14 & 43.25 $\pm$ 0.14 & 22.58 $\pm$ 0.40 \\ 
     & MACER \cite{Zhai2020MACER} & 91.59 $\pm$ 0.20 & 83.44 $\pm$ 0.35 & 71.10 $\pm$ 0.45 & 55.67 $\pm$ 0.27 & 38.67 $\pm$ 0.33 & 20.09 $\pm$ 0.64 \\ 
     & Consistency \cite{jeong2020consistency} & 94.96 $\pm$ 0.02 & 89.75 $\pm$ 0.07 & 79.70 $\pm$ 0.09 & 63.54 $\pm$ 0.12 & 41.74 $\pm$ 0.13 & 20.22 $\pm$ 0.25 \\ 
    \cmidrule(l){2-2} \cmidrule(l){3-8}
    & \textbf{SmoothMix} ($\eta$ = 1.0) & 95.52 $\pm$ 0.08 & 90.50 $\pm$ 0.07 & \textbf{80.55 $\pm$ 0.09} & \textbf{64.09 $\pm$ 0.15} & \textbf{43.16 $\pm$ 0.05} & \textbf{\underline{23.94} $\pm$ 0.19} \\ 
    & \textbf{+ One-Step adversary} & 94.72 $\pm$ 0.07 & 89.40 $\pm$ 0.09 & 79.46 $\pm$ 0.08 & \textbf{64.04 $\pm$ 0.08} & \textbf{\underline{44.82} $\pm$ 0.09} & \textbf{\underline{27.35} $\pm$ 0.15} \\ 
    & \textbf{SmoothMix} ($\eta$ = 5.0) & 93.71 $\pm$ 0.04 & 88.00 $\pm$ 0.05 & 77.95 $\pm$ 0.13 & \textbf{62.78 $\pm$ 0.08} & \textbf{\underline{44.87} $\pm$ 0.14} & \textbf{\underline{28.88} $\pm$ 0.16} \\ 
    & \textbf{+ One-Step adversary} & 93.11 $\pm$ 0.05 & 87.24 $\pm$ 0.07 & 77.22 $\pm$ 0.10 & \textbf{62.48 $\pm$ 0.15} & \textbf{\underline{44.85} $\pm$ 0.05} & \textbf{\underline{29.66} $\pm$ 0.15} \\ 
    \bottomrule
\end{tabular}

        \end{adjustbox}
    \end{table*}
    
    \begin{table}[ht]
    \centering
    \small
        \caption{Comparison of ACR for various training methods on MNIST. The reported values are the mean and standard deviation across 5 seeds. We set our result bold-faced whenever the value improves the Gaussian baseline, and the underlined are best-performing model per $\sigma$.}
	    \label{tab:mnist_var}
        \begin{tabular}{lccc}
        \toprule
        ACR (MNIST) & $\sigma=0.25$ & $\sigma=0.50$ & $\sigma=1.00$  \\ 
        \midrule
        Gaussian \cite{pmlr-v97-cohen19c} & 0.9108\pms{0.0003} & 1.5581\pms{0.0016} & 1.6184\pms{0.0021} \\
        Stability \cite{li2019stab} & 0.9152\pms{0.0007} & 1.5719\pms{0.0028} & 1.6341\pms{0.0018}  \\ 
        SmoothAdv \cite{nips_salman19} & {0.9322}\pms{0.0005} & 1.6872\pms{0.0007} & 1.7786\pms{0.0017}  \\
        MACER  \cite{Zhai2020MACER} & 0.9201\pms{0.0006} & 1.5899\pms{0.0069} & 1.5950\pms{0.0051}  \\ 
        Consistency \cite{jeong2020consistency} & {0.9279\pms{0.0003}}  & {1.6549\pms{0.0011}} & {1.7376\pms{0.0017}}  \\
        \cmidrule(l){1-1} \cmidrule(l){2-4}
        \textbf{SmoothMix ($\eta=1.0$)} & \textbf{0.9296\pms{0.0003}} & \textbf{1.6776\pms{0.0007}} & \textbf{1.7867\pms{0.0020}} \\
        \textbf{+ One-Step adversary} & \textbf{0.9330\pms{0.0004}} & {\textbf{\underline{1.6932}\pms{0.0009}}} & \textbf{1.8169\pms{0.0011}} \\
        \textbf{SmoothMix ($\eta=5.0$)} & \textbf{0.9317\pms{0.0002}} & {\textbf{\underline{1.6932}\pms{0.0007}}} & \textbf{1.8185\pms{0.0016}} \\
        \textbf{+ One-Step adversary} & {\textbf{\underline{0.9332}\pms{0.0002}}} & \textbf{1.6851\pms{0.0003}} & {\textbf{\underline{1.8212}\pms{0.0013}}} \\
        \bottomrule
        \end{tabular}
    \end{table}

\FloatBarrier
\section{Additional results on CIFAR-10}
\label{ap:add_cifar10}

In this section, we report additional experimental results on CIFAR-10 \cite{dataset/cifar}, namely with $\sigma=1.0$ (see Table~\ref{tab:cifar10} for the results for $\sigma\in\{0.25, 0.5\}$). We follow the same experimental details as specified in Section~\ref{ss:result_cifar10} and Appendix~\ref{ap:details}, including the common hyperparameter choice of $\eta=5.0$ for SmoothMix for other experiments as well. Again, we compare our method with various existing robust training methods for smoothed classifiers \cite{pmlr-v97-cohen19c, li2019stab, nips_salman19, Zhai2020MACER, jeong2020consistency}, and Table~\ref{tab:add_cifar10} summarizes the results. Overall, we still observe a similar trend to Section~\ref{ss:result_cifar10} that (a) ``SmoothMix'' offers a significant improvement of robust accuracy without compromising the clean accuracy much, and (b) incorporating the one-step adversary thus can further complementarily boost ACR to outperform other state-of-the-art baseline training methods: \eg it is notable that ``SmoothMix + One-step adversary'' achieves fairly comparable or better robust accuracy than MACER while maintaining much higher clean accuracy, \ie the certified test accuracy at $r=0.0$, namely 41.4 $\rightarrow$ 45.1. This confirms that our proposed SmoothMix can offer a better trade-off between accuracy and certified robustness during training.

\begin{table*}[ht]
\centering
\caption{Comparison of approximate certified test accuracy (\%) and ACR on CIFAR-10. All the models are trained and evaluated with the same smoothing factor specified by $\sigma$. Each value except ACR indicates the fraction of test samples those have $\ell_2$ certified radius larger than the threshold specified at the top row. We set our results bold-faced whenever the value improves the Gaussian baseline, and underlined whenever the value achieves the best among the considered baselines. $^*$~indicates that the results are evaluated from the official pre-trained models released by authors.}
\label{tab:add_cifar10}
    \vspace{0.03in}
    \begin{adjustbox}{width=1\linewidth}
    \begin{tabular}{clc|cccccccccc}
    \toprule
    $\sigma$ &  Models (CIFAR-10) & ACR & 0.00 & 0.25 & 0.50 & 0.75 & 1.00 & 1.25 & 1.50 & 1.75 & 2.00 & 2.25 \\ 
    \midrule
    \multirow{7.5}{*}{1.00}& Gaussian \cite{pmlr-v97-cohen19c} & 0.542 & 47.2 & 39.2 & 34.0 & 27.8 & 21.6 & 17.4 & 14.0 & 11.8 & 10.0 & 7.6 \\
    & Stability training \cite{li2019stab} & 0.526  & 43.5 & 38.9 & 32.8 & 27.0 & 23.1 & 19.1 & 15.4 & 11.3 & 7.8 & 5.7 \\
    & SmoothAdv$^*$ \cite{nips_salman19} & 0.660 & 50.8 & 44.9 & 39.0 & 33.6 & 28.5 & 23.7 & 19.4 & 15.4 & 12.0 & 8.7  \\ 
    & MACER$^*$  \cite{Zhai2020MACER} & {0.744} & {41.4} & {38.5} & {35.2} & {32.3} & {29.3} & {26.4} & {23.4} & {20.2} & {17.4} & {14.5}  \\
    & {Consistency \cite{jeong2020consistency}}  & {{0.756}} & 46.3 & {42.2} & {38.1} & {34.3} & {30.0} & {26.3} & {22.9} & {19.7} & {16.6} & {13.8} \\
    \cmidrule(l){2-2} \cmidrule(l){3-3} \cmidrule(l){4-13}
    & \textbf{SmoothMix (Ours)}  & \textbf{0.725} & 47.1 & \textbf{42.5} & \textbf{37.5} & \textbf{32.9} & \textbf{28.7} & \textbf{24.9} & \textbf{21.3} & \textbf{18.3} & \textbf{15.5} & \textbf{12.6} \\
    & \textbf{+ One-step adversary}  & \underline{\textbf{0.773}} & 45.1 & \textbf{41.5} & \textbf{37.5} & \textbf{33.8} & \underline{\textbf{30.2}} & \underline{\textbf{26.7}} & \underline{\textbf{23.4}} & \underline{\textbf{20.2}} & \textbf{17.2} & \underline{\textbf{14.7}} \\
    \bottomrule
\end{tabular}
    \end{adjustbox}
\end{table*}

\FloatBarrier
\section{Results on ImageNet}
\label{ap:imagenet}

We also compare our method on ImageNet \cite{dataset/ilsvrc} classification dataset, to verify the scalability of the method on large-scale datasets. In this experiment, we perform our evaluation on the sub-sampled validation dataset of ImageNet with 500 samples following the previous works \cite{pmlr-v97-cohen19c, nips_salman19, jeong2020consistency}.
When SmoothMix is used, we simply set $T=1$ and $m=1$ mainly in order to reduce the overall training cost, and we fix $\alpha=8.0$ for both cases of $\sigma=0.5, 1.0$: this choice leads larger $\alpha\cdot T$ when $\sigma=0.5$ compared to the MNIST and CIFAR-10 experiments, but we empirically observe that ImageNet is less sensitive to $\alpha\cdot T$, possibly due to that ImageNet consists of higher-resolution inputs, \ie higher input dimension accordingly, than the others.
We use the one-step adversary (Section~\ref{ss:smoothmix}) by default here, but we make sure that each adversarial example (found with a large $\alpha$) is further projected in a $\ell_2$-ball of $\epsilon=1.0$ before it replaces the clean sample, which can be done without adding significant computational overhead.
Table~\ref{tab:imagenet} summarizes the results, and we still observe the effectiveness of SmoothMix compared to the baseline methods, both in terms of ACR and certified test accuracy.

\begin{table}[ht]
\centering
\caption{Comparison of approximate certified test accuracy (\%) on ImageNet. We set our results bold-faced whenever the value improves the Gaussian baseline, and underlined whenever the value achieves the best among the considered baselines.}
\label{tab:imagenet}
\vspace{0.03in}
\small
    \begin{tabular}{clccccccccc}
    \toprule
    $\sigma$ &  Models (ImageNet) & ACR & 0.0 & 0.5 & 1.0 & 1.5 & 2.0 & 2.5 & 3.0 & 3.5 \\
    \midrule
    \multirow{4.5}{*}{0.50}& Gaussian \cite{pmlr-v97-cohen19c} & 0.733 & 57 & 46 & 37 & 29 & 0 & 0 & 0 & 0  \\
    & Consistency \cite{jeong2020consistency}  & {0.822} & 55 & {50} & {44} & {34} & 0 & 0 & 0 & 0 \\
    & SmoothAdv \cite{nips_salman19} & {0.825} & 54 & {49} & {43} & {37} & 0 & 0 & 0 & 0  \\
    \cmidrule(l){2-2} \cmidrule(l){3-3} \cmidrule(l){4-11}
    & \textbf{SmoothMix (Ours)} & \underline{\textbf{0.846}} & 55 & \underline{\textbf{50}} & {\textbf{43}} & \underline{\textbf{38}} & 0 & 0 & 0 & 0 \\
    \midrule
    \multirow{4.5}{*}{1.00}& Gaussian \cite{pmlr-v97-cohen19c} & 0.875 & 44 & 38 & 33 & 26 & 19 & 15 & 12 & 9  \\
    & Consistency \cite{jeong2020consistency}  & {0.982} & 41 & 37 & 32 & {28} & {24} & {21} & {17} & {14} \\
    & SmoothAdv \cite{nips_salman19} & {1.040} & 40 & 37 & {34} & {30} & {{27}} & {{25}} & 20 & 15 \\
    \cmidrule(l){2-2} \cmidrule(l){3-3} \cmidrule(l){4-11}
    & \textbf{SmoothMix (Ours)} & \underline{\textbf{1.047}} & 40 & 37 & \underline{\textbf{34}} & \underline{\textbf{30}} & \textbf{26} & \textbf{24} & \underline{\textbf{20}} & \underline{\textbf{17}} \\
    \bottomrule
    \end{tabular}
\end{table}

\FloatBarrier
\section{Detailed results on ablation study}
\label{ap:ablation}

In this section, we report the detailed numerical results and more discussions on the ablation study presented in Section~\ref{ss:ablation}. Here, Table~\ref{tab:ab_eta}, \ref{tab:ab_aT}, \ref{tab:ab_maxnorm} and \ref{tab:ab_m} presented in what follow are the detailed results for Figure~\ref{fig:eta}, \ref{fig:alpha_steps}, \ref{fig:maxnorm} and \ref{fig:noise_vecs}, respectively. 

\begin{table*}[ht]
\centering
\caption{Comparison of ACR and approximate certified test accuracy on MNIST for varying $\eta$. We assume $\sigma=1.0$ in this experiment. “Gaussian” indicates the baseline training with Gaussian augmentation. We set the results bold-faced whenever the value improves ``Gaussian''.}
\label{tab:ab_eta}
    \vspace{0.03in}
\begin{adjustbox}{width=1\linewidth}
\begin{tabular}{cc|ccccccccccc}
    \toprule
    Setups & ACR & 0.00 & 0.25 & 0.50 & 0.75 & 1.00 & 1.25 & 1.50 & 1.75 & 2.00 & 2.25 & 2.50 \\ 
    \midrule
     Gaussian & 1.620 & 96.4 & 94.4 & 91.4 & 87.0 & 79.9 & 71.0 & 59.6 & 46.2 & 32.6 & 19.7 & 10.8  \\
    \midrule
    $\eta=\pz1$ & \textbf{1.789} & 95.5 & 93.6 & 90.5 & 86.2 & \textbf{80.7} & \textbf{73.7} & \textbf{64.1} & \textbf{53.9} & \textbf{43.1} & \textbf{33.5} & \textbf{24.1} \\
    $\eta=\pz2$ & \textbf{1.810} & 94.9 & 92.7 & 89.7 & 85.1 & 79.6 & \textbf{72.6} & \textbf{63.8} & \textbf{54.0} & \textbf{44.4} & \textbf{35.4} & \textbf{26.6} \\
    $\eta=\pz4$ & \textbf{1.820} & 94.0 & 91.8 & 88.4 & 83.9 & 78.3 & \textbf{71.4} & \textbf{63.0} & \textbf{53.6} & \textbf{44.9} & \textbf{36.8} & \textbf{28.7} \\
    $\eta=\pz8$ & \textbf{1.817} & 93.4 & 91.0 & 87.5 & 82.7 & 77.3 & 70.2 & \textbf{62.4} & \textbf{53.0} & \textbf{44.8} & \textbf{37.0} & \textbf{29.3} \\
    $\eta=16$ & \textbf{1.812} & 92.9 & 90.3 & 86.7 & 82.1 & 76.6 & 69.7 & \textbf{61.8} & \textbf{52.6} & \textbf{44.5} & \textbf{36.9} & \textbf{29.6} \\
    \bottomrule
\end{tabular}
\end{adjustbox}
\end{table*}

\begin{table*}[ht]
\centering
    \caption{Comparison of ACR and approximate certified test accuracy on MNIST for varying $\alpha$ and $T$ under control of $\alpha\cdot T=8$. We assume $\sigma=1.0$ in this experiment. “Gaussian” indicates the baseline training with Gaussian augmentation. We set the results bold-faced whenever the value improves ``Gaussian''.}
\label{tab:ab_aT}
    \vspace{0.03in}
\begin{adjustbox}{width=1\linewidth}
\begin{tabular}{cc|ccccccccccc}
    \toprule
    Setups & ACR & 0.00 & 0.25 & 0.50 & 0.75 & 1.00 & 1.25 & 1.50 & 1.75 & 2.00 & 2.25 & 2.50 \\ 
    \midrule
     Gaussian & 1.620 & 96.4 & 94.4 & 91.4 & 87.0 & 79.9 & 71.0 & 59.6 & 46.2 & 32.6 & 19.7 & 10.8  \\
    \midrule
    $(\alpha, T)=(8.0, 1)$ & \textbf{1.785} & 95.5 & 93.5 & 90.5 & 86.0 & \textbf{80.5} & \textbf{73.1} & \textbf{63.9} & \textbf{53.5} & \textbf{43.3} & \textbf{33.2} & \textbf{24.0} \\ 
    $(\alpha, T)=(4.0, 2)$ & \textbf{1.788} & 95.4 & 93.4 & 90.4 & 85.9 & \textbf{80.5} & \textbf{73.5} & \textbf{63.9} & \textbf{53.5} & \textbf{43.1} & \textbf{33.4} & \textbf{24.4} \\
    $(\alpha, T)=(2.0, 4)$ & \textbf{1.790} & 95.5 & 93.5 & 90.7 & 86.2 & \textbf{80.7} & \textbf{73.7} & \textbf{64.3} & \textbf{53.9} & \textbf{43.2} & \textbf{33.4} & \textbf{23.8} \\
    $(\alpha, T)=(1.0, 8)$ & \textbf{1.789} & 95.5 & 93.6 & 90.5 & 86.2 & \textbf{80.7} & \textbf{73.7} & \textbf{64.1} & \textbf{53.9} & \textbf{43.1} & \textbf{33.5} & \textbf{24.1} \\
    \bottomrule
\end{tabular}
\end{adjustbox}
\end{table*}

\begin{table*}[ht]
\centering
\caption{Comparison of ACR and approximate certified test accuracy on MNIST for varying $\varepsilon$, the hard limit on $\ell_2$-norm of adversarial perturbations. We assume $\sigma=1.0$ in this experiment. “Gaussian” indicates the baseline training with Gaussian augmentation. ``$\varepsilon=\infty$'' denotes our original setup of unrestricted adversarial attacks. We set the results bold-faced whenever the value improves ``Gaussian''.}
\label{tab:ab_maxnorm}
    \vspace{0.03in}
\begin{adjustbox}{width=1\linewidth}
\begin{tabular}{cc|ccccccccccc}
    \toprule
    Setups & ACR & 0.00 & 0.25 & 0.50 & 0.75 & 1.00 & 1.25 & 1.50 & 1.75 & 2.00 & 2.25 & 2.50 \\ 
    \midrule
     Gaussian & 1.620 & 96.4 & 94.4 & 91.4 & 87.0 & 79.9 & 71.0 & 59.6 & 46.2 & 32.6 & 19.7 & 10.8  \\
    \midrule
    $\varepsilon=2.0$ & \textbf{1.723} & 96.1 & 94.3 & 91.4 & \textbf{87.1} & \textbf{81.2} & \textbf{73.6} & \textbf{63.7} & \textbf{52.1} & \textbf{39.8} & \textbf{28.2} & \textbf{16.6} \\
    $\varepsilon=4.0$ & \textbf{1.751} & 95.9 & 94.0 & 91.1 & 86.8 & \textbf{81.0} & \textbf{73.7} & \textbf{64.3} & \textbf{53.1} & \textbf{41.4} & \textbf{30.6} & \textbf{19.8} \\
    $\varepsilon=6.0$ & \textbf{1.778} & 95.6 & 93.7 & 90.6 & 86.5 & \textbf{80.8} & \textbf{73.7} & \textbf{64.4} & \textbf{53.8} & \textbf{42.8} & \textbf{32.6} & \textbf{22.9} \\
    $\varepsilon=8.0$ & \textbf{1.788} & 95.5 & 93.5 & 90.4 & 86.1 & \textbf{80.5} & \textbf{73.5} & \textbf{64.2} & \textbf{53.8} & \textbf{43.2} & \textbf{33.5} & \textbf{24.1} \\
    \midrule
    $\varepsilon=\infty$ (Ours) & \textbf{1.789} & 95.5 & 93.6 & 90.5 & 86.2 & \textbf{80.7} & \textbf{73.7} & \textbf{64.1} & \textbf{53.9} & \textbf{43.1} & \textbf{33.5} & \textbf{24.1}\\
    \bottomrule
\end{tabular}
\end{adjustbox}
\end{table*}

\begin{table*}[ht]
\centering
\caption{Comparison of ACR and approximate certified test accuracy on MNIST for varying $m$, the number of noise samples used for estimating smoothed predictions. We assume $\sigma=1.0$ in this experiment. “Gaussian” indicates the baseline training with Gaussian augmentation. We set the results bold-faced whenever the value improves ``Gaussian''.}
\label{tab:ab_m}
    \vspace{0.03in}
\begin{adjustbox}{width=1\linewidth}
\begin{tabular}{cc|ccccccccccc}
    \toprule
    Setups & ACR & 0.00 & 0.25 & 0.50 & 0.75 & 1.00 & 1.25 & 1.50 & 1.75 & 2.00 & 2.25 & 2.50 \\ 
    \midrule
     Gaussian & 1.620 & 96.4 & 94.4 & 91.4 & 87.0 & 79.9 & 71.0 & 59.6 & 46.2 & 32.6 & 19.7 & 10.8  \\
    \midrule
    $m=1$ & \textbf{1.744} & 94.5 & 92.2 & 88.9 & 84.1 & 78.1 & 70.9 & \textbf{61.9} & \textbf{51.7} & \textbf{41.7} & \textbf{31.9} & \textbf{23.2} \\
    $m=2$ & \textbf{1.776} & 95.3 & 93.0 & 89.8 & 85.4 & 79.8 & \textbf{72.7} & \textbf{63.5} & \textbf{53.1} & \textbf{42.6} & \textbf{33.0} & \textbf{24.0} \\
    $m=4$ & \textbf{1.789} & 95.5 & 93.6 & 90.5 & 86.2 & \textbf{80.7} & \textbf{73.7} & \textbf{64.1} & \textbf{53.9} & \textbf{43.1} & \textbf{33.5} & \textbf{24.1} \\
    $m=8$ & \textbf{1.788} & 95.9 & 93.9 & 91.0 & 86.7 & \textbf{81.0} & \textbf{73.9} & \textbf{64.6} & \textbf{54.1} & \textbf{43.2} & \textbf{33.1} & \textbf{23.3} \\
    \bottomrule
\end{tabular}
\end{adjustbox}
\end{table*}

\FloatBarrier

\end{document}